\documentclass[twoside,11pt]{article}

\usepackage{jmlr2e}
\usepackage{amsmath,amsfonts,amssymb,amsbsy}
\usepackage{bbold}
\usepackage{dsfont} %
\usepackage{url}
\usepackage{graphicx}
\usepackage{array}
\usepackage{hhline}
\usepackage{dcolumn}
\usepackage{footnote}
\usepackage{placeins} %
\usepackage{tikz,pgfplots}
\usepackage[titletoc]{appendix} %
\usepackage{natbib} %
\usepackage{booktabs} %
\usepackage{soul} %
\usepackage{times}      %
\usepackage{verbatim}
\usepackage{microtype}

\pdfoutput=1
\usepackage{hyperref}
\hypersetup{
     colorlinks=true,
     linkcolor=black,
     citecolor=blue,
     filecolor=black,
     urlcolor=black,
}

\usepackage{ifthen}
\usetikzlibrary{matrix}
\usetikzlibrary{calc}

\def\figpdfdir{fig/} %
\def\figtikzdir{tikz/} %

\newcommand{\minput}[2][]{
\ifthenelse{\equal{#1}{pdf}}
	{ \includegraphics{\figpdfdir #2} }
	{ \tikzsetnextfilename{#2} \input{\figtikzdir #2} } 
}

\usetikzlibrary{external}
\tikzset{external/system call={lualatex
	\tikzexternalcheckshellescape -halt-on-error -interaction=batchmode
	-jobname "\image" "\texsource"}}
\newcommand{\vc}[1] { \mathbf{#1} }
\newcommand{\vs}[1] { \boldsymbol{#1} }
\newcommand{\tp}{\mathsf{T}}
\newcommand{\ti}[1] { \tilde{#1} } 
\newcommand{\mc}[1] { \mathcal{#1} } 
\newcommand{\tx}[1] { \text{#1} } 
\newcommand{\given} { \,|\, }
\newcommand{\data} {\mc D}
\renewcommand{\Re} {\mathds{R}}
\newcommand{\indicator}[1] { \mathds{1} \left(#1\right) } %

\newcommand{\mean}[2][] {  \mathrm{E}_{#1} {\left( \, #2 \, \right)}  }
\newcommand{\var}[1] { \mathrm{Var} {\left(#1\right)} }
\newcommand{\svar}[3][] { \mathrm{V}_{#1}^{#2} {\left[#3\right]} } 
\newcommand{\cov}[1] { \mathrm{Cov} {\left(#1\right)} }
\newcommand{\cor}[1] { \mathrm{Cor} {\left(#1\right)} }
\newcommand{\tr}[1] { \mathrm{Tr} {\left(#1\right)} }
\newcommand{\KL}[2] { \mathrm{KL} {\left(#1 \, \| \, #2\right)} }

\newcommand{\Unif}[1]{ \mathrm{U} {\left(#1\right)} }
\newcommand{\Normal}[1] { \mathrm{N} {\left(#1\right)}  }

\newcommand{\halfStudent}[2][] { t_{#1}^+ {\left(#2\right)} }
\newcommand{\InvChi}[1] { \mathrm{\text{-}Inv\text{-}}\chi^2 {\left(#1\right)} }

\newcommand{\halfCauchy}[1] { \mathrm{C}^+ {\left(#1\right)} }

\newcommand{\Ber}[1] {\mathrm{Ber}{\left(#1\right)} }

\newcommand{\captionspace}{\vspace{-0.5cm}}

\jmlrheading{1}{2018}{0-0}{00/00}{00/00}{meila00a}{Juho Piironen, Markus Paasiniemi and Aki Vehtari}
\ShortHeadings{Projection predictive inference}{Piironen, Paasiniemi and Vehtari}
\firstpageno{1}

\makeatletter
\def\@starteditor{\noindent \small {}}
\makeatother

\begin{document}

\title{Projective Inference in High-dimensional Problems:\\ Prediction and Feature Selection}
\author{\name Juho Piironen \email juho.piironen@aalto.fi\\
		\name Markus Paasiniemi \email markus.paasiniemi@helsinki.fi \\
		\name Aki Vehtari \email aki.vehtari@aalto.fi\\
		\addr Helsinki Institute of Information Technology (HIIT) \\
		Department of Computer Science, Aalto University \\
		P.O.Box 15400, FI-00076 Aalto, Finland
}

\editor{}

\maketitle
\thispagestyle{plain} 

\begin{abstract}%
This paper discusses predictive inference and feature selection for generalized linear models with scarce but high-dimensional data.
We argue that in many cases one can benefit from a decision theoretically justified two-stage approach: first, construct a possibly non-sparse model that predicts well, and then find a minimal subset of features that characterize the predictions.
The model built in the first step is referred to as the \emph{reference model} and the operation during the latter step as predictive \emph{projection}.
The key characteristic of this approach is that it finds an excellent tradeoff between sparsity and predictive accuracy, and the gain comes from utilizing all available information including prior and that coming from the left out features.
We review several methods that follow this principle and provide novel methodological contributions.
We present a new projection technique that unifies two existing techniques and is both accurate and fast to compute.
We also propose a way of evaluating the feature selection process using fast leave-one-out cross-validation that allows for easy and intuitive model size selection.
Furthermore, we prove a theorem that helps to understand the conditions under which the projective approach could be beneficial.
The benefits are illustrated via several simulated and real world examples.
\end{abstract}
\vspace{0.5\baselineskip}


\section{Introduction}

Predictive inference and feature selection for generalized linear models (GLMs) in problems with scarce data but high-dimensional feature space---regime known as ``small $n$, large $p$''\,\footnote{Due to this historical naming we stick with these symbols but also use $p$ to denote density functions. We hope this does not confuse the reader.}---remains a topic of active research.
Often, albeit not always, the goals are twofold: the desire is to find a model that predicts unseen data well but utilizes only a small subset of features thereby facilitating the interpretation and making the model more convenient to use at prediction time.

A vast variety of different approaches have been proposed.
Frequentist approaches typically formulate an estimator with a penalty that enforces sparsity in the solution \citep[e.g.,][]{breiman1995,tibshirani1996,fan2001,zou2005,candes2007}.
A useful overview has been written by~\cite{hastie2015book}.
Among Bayesians, the most common approach is to use a sparsifying prior that favors solutions with a small number of active predictors \citep[e.g.,][]{george1993,raftery1997,ishwaran2005,johnson2012,carvalho2010}.
These approaches do not automatically produce truly sparse solutions since there is always nonzero probability for each feature being included in the model, but sparse models can be obtained for instance by removing features with estimated posterior effect below certain threshold \citep{barbieri2004,ishwaran2005,narisetty2014}.

All these approaches attempt to solve the two problems---prediction and feature selection---simultaneously.
In this paper we argue that in many situations one can gain if these problems are solved in two stages, by first finding a model that predicts well (not caring about true sparsity) and then finding a minimal subset of features that provide similar predictions as this model which we shall call a \emph{reference model}.
This strategy not only solves many issues that one might encounter in traditionally used Bayesian approaches (as we will discuss in Sec.~\ref{sec:traditional_approaches}) but has also shown empirically very good performance in comparison to many other methods with good tradeoff between sparsity and predictive accuracy~\citep{piironen2017a}.
Our discussion will be mainly from the Bayesian viewpoint but is aimed to provide several insights also for a non-Bayesian oriented reader since the reference model approach is not intrinsically limited only to the Bayesian paradigm.

A piece of pioneering work in this line was carried out by~\cite{lindley1968}, who considered prediction in linear regression model when some of the features are unavailable at prediction time.
A related but slightly different approach was proposed by~\cite{goutis1998} and~\cite{dupuis2003} who introduced the concept of \emph{projecting} the posterior information in the reference model to smaller submodels, although they were mainly interested in feature selection and less so about predicting with the submodels.
A few related papers are due to \cite{nott2010} and \cite{tran2012}, who introduced variants of the original projection of Goutis, Dupuis and Robert.
These ideas are also very closely related to the frequentist technique known as ``preconditioning'' for feature selection~\citep{paul2008}. 
Also the approach of \cite{hahn2015} utilizes essentially the same idea but formulates the projection in a slightly different way.
It is worth pointing out that the same conceptual idea of replacing a large complex model with a smaller one has also successfully been used in the neural networks literature---where it is known as ``model compression'' or ``knowledge distillation''---although there the interests are reduced memory costs and faster out-of-sample predictions rather than feature selection~\citep{bucila2006,hinton2015}. 

In general we assume that the phenomena we are modeling are so complex that the true model is not included in the list of models under consideration.
Hence we adopt the $\mc M$-open view although the projective approach is partially using also the $\mc M$-completed view.\footnote{As there are alternative definitions of $\mc M$-closed, -open, and -completed, we emphasize that we use definitions as in \cite{bernardo1994book} and \cite{vehtari2012}.}  
More specifically, in the first stage where we attempt to construct a model that predicts well, we assume $\mc M$-open case.
If we are able to construct a sensible model that passes model assessment and checking \citep[see, e.g.,][]{gelman2013book,gabry2018}, we can use that model as the reference model in $\mc M$-completed setting. 
We assume the reference model is our best description of the future data, and in $\mc M$-completed setting the predictive performances of other models are evaluated with respect to that reference model. 
The use of a reference model reduces the variance in the model selection in a same way as use of model assumptions reduce the uncertainty in the usual data modeling. 
Finally, as we do select a single model, we estimate the effect of selection process with $\mc M$-open style using cross-validation (see Section~\ref{sec:validation}).

\subsection{Our contributions}

This paper makes several contributions which are summarized as follows:
\begin{itemize}
	\item We review the aforementioned projection techniques under unified notation, illustrate their differences in detail and give recommendations about the preferred approaches depending on the problem at hand.
	\item We develop a new type of projection---called clustered projection---that can be considered as a unification of the approach of Goutis, Dupuis and Robert and that of Tran et al., and show that it gives a good balance between speed and accuracy.
	\item We propose a new efficient method for validating the selection process using approximate leave-one-out (LOO) cross-validation.
	This technique can be used to assess the predictive accuracy of the submodels which allows for intuitive model size selection.
	\item We discuss the typical difficulties encountered with the traditional Bayesian approaches via small examples and show how the projective approach yields much more satisfactory results. 
	Since an extensive comparison showing the superiority of the projection (in terms of sparsity-accuracy tradeoff) to many other Bayesian model selection strategies over a variety of data sets has already been carried out earlier~\citep{piironen2017a}, here we focus only on some of the most commonly used techniques and illustrate via small examples \emph{why} they are problematic.
	\item We discuss the connection of the projection to the popular Lasso estimator~\citep{tibshirani1996} in detail together with several empirical results that demonstrate the benefit of the proposed approach in the ``small $n$, large $p$'' -	setting.
	\item We prove a theorem that---at least in our knowledge---for the first time gives a theoretical argument of why and under which conditions the use of reference model could be beneficial for parameter learning in linear models.
	\item We provide an R software package {\tt projpred} that implements all the discussed methods. The package is freely available and makes the method easily accessible to a wide audience.\,\footnote{The codes with installation instructions and examples are available at~\url{https://github.com/stan-dev/projpred}.}
\end{itemize}

\subsection{Why does a reference model improve feature selection?}
\label{sec:intro_example}

We begin with a simple example that motivates why use of a reference model can be useful for feature selection.
Although the details are different, this example is greatly inspired by the one presented by \cite{paul2008}.

Assume we have collected $n$ measurements of $p$ features $x_j$,\, $j=1,\dots,p$ along with measurements of some target variable $y$. 
Assume also that the data are generated according to the following mechanism:
\begin{equation}
\begin{alignedat}{2}
	f &\sim \Normal{0,1}, &&  \\
	y \given f &\sim \Normal{f, 1} && \\
	x_j \given f &\sim \Normal{\sqrt{\rho}f,\, 1-\rho},  \qquad && j = 1,\dots,p_\tx{rel} \,, \\
	x_j \given f &\sim \Normal{0, 1}, && j = p_\tx{rel}+1,\dots,p \,.
\end{alignedat}
\label{eq:toy_data}
\end{equation}
The target variable values $y$ are noisy observations from the latent function values $f$ which are drawn randomly from a standard Gaussian distribution.
The first $p_\tx{rel}$ features~$x_j$ are also noisy observations from the latent function $f$, which makes them correlated and on average equally predictive about~$y$.
The multiplier $\sqrt{\rho}$ and the noise variance $1-\rho$ are chosen so that the marginal variance of each~$x_j$ is~$1$ and the pairwise correlations between the first $p_\tx{rel}$ features are all equal to~$\rho$.
The rest of the features are drawn randomly from a standard normal distribution and are thus uncorrelated and irrelevant for predicting $y$.

Suppose our goal is to assess how predictive each of the features is about the target variable.
A simple strategy would be to compute the sample correlation $R(x_j,y)$ between each feature and the target variable and then rank the features based on the absolute values $|R(x_j,y)|$.
Since the features are related to the target variable via the latent $f$, clearly our task would be easier if we had access to the noiseless values $f$ instead of the noisy ones~$y$, since the additional noise weakens the correlations, that is, $|\cor{x_j,y}| < |\cor{x_j,f}|$ for $j=1,\dots,p_\tx{rel}$. 
In practice we do not observe $f$ directly, but intuitively if we could build up a model whose output $f_*$ is fairly close to the true~$f$, we might expect to benefit by making the assessment based on the sample correlations $R(x_j,f_*)$ instead of $R(x_j,y)$.

\begin{figure}%
	\centering
	\minput[pdf]{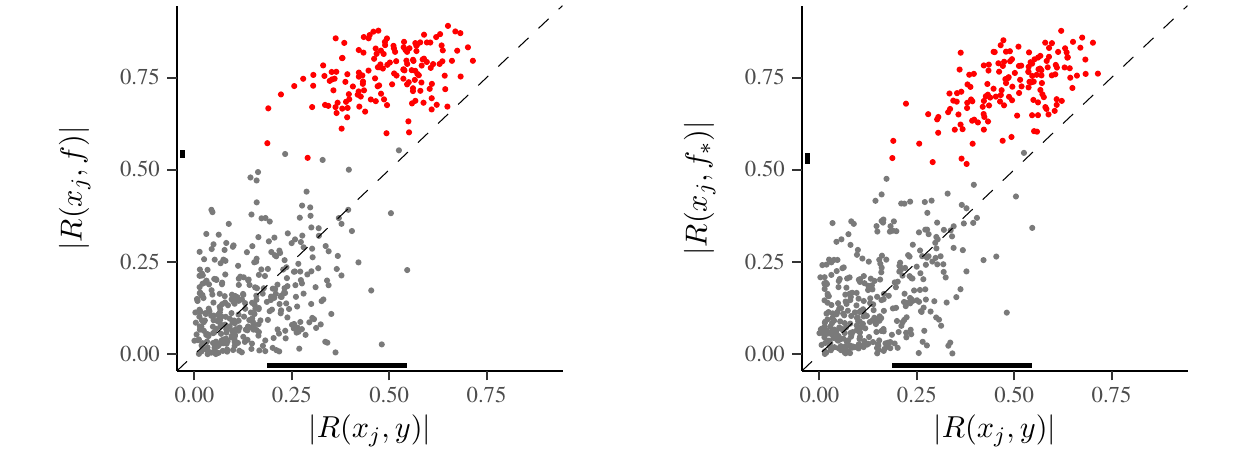}
	\caption{ {\it Introductory example:} Left: Absolute sample correlations of each feature $x_j$ with the observed target variable $y$ (horizontal axis) and with the noiseless latent value $f$ (vertical axis) for $n=30$ observations generated according to~\eqref{eq:toy_data}, with $p=500$, $p_\tx{rel}=150$ and $\rho=0.5$. Red dots denote the truly relevant features and gray dots irrelevant noise features. Right: The same but the true latent $f$ replaced by the predictions $f_*$ of a reference model we can actually compute (see the text for details). The relevant features are much better separated from the irrelevant ones when we consider their correlations with either the true~$f$ or the reference model predictions $f_*$ instead of the observed $y$ (the amount of overlap between the two groups is depicted by the black lines). }
	\label{fig:univariate_example}
\end{figure}

Figure~\ref{fig:univariate_example} illustrates this idea.
The left graph shows the absolute sample correlations $|R(x_j,y)|$ versus $|R(x_j,f)|$ for one data realization from~\eqref{eq:toy_data} with $p=500$, $p_\tx{rel}=150$, $n=30$ and $\rho=0.5$.
The relevant features (red dots) are much better separated from the irrelevant ones (gray dots) when we consider their correlation with $f$ instead of $y$.
The right graph demonstrates that this holds also when we replace the unknown $f$ with predictions $f_*$ of a reference model we can actually compute.
Here the reference fit is obtained by Bayesian linear regression of $y$ on the first three supervised principal components of all the features (the procedure is discussed in detail in Sec.~\ref{sec:refmodel_construction}).

\begin{figure}%
	\centering
	\minput[pdf]{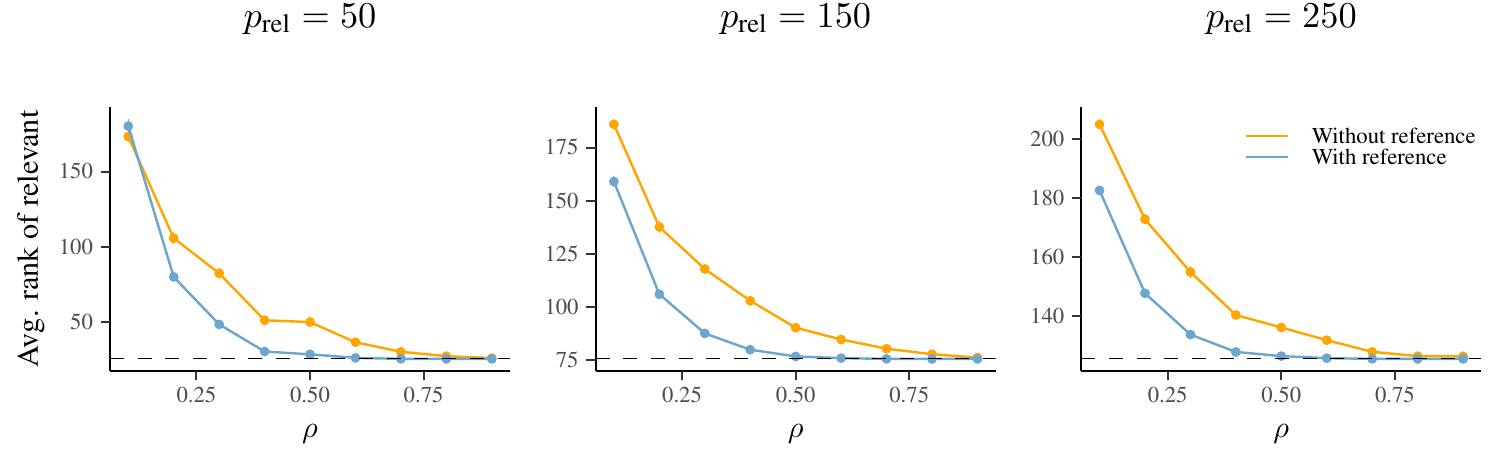}
	\captionspace
	\caption{ {\it Introductory example:} Average rank of the truly relevant features when the features are sorted based on their absolute sample correlations with $y$ (orange) or with the reference model predictions $f_*$ (blue). The results are averages over 100 data realizations from mechanism~\eqref{eq:toy_data}, with $n=30$ and $p=500$, and the results are shown for three different values of~$p_\tx{rel}$ with varying $\rho$. Lower values are better and the dashed lines denote the oracle results (that is, if all truly relevant features are ranked before the irrelevant ones). The standard errors (vertical lines) are in most cases smaller than the dot sizes.}
	\label{fig:univariate_rank}
\end{figure}

Figure~\ref{fig:univariate_rank} shows that this pattern holds for a wide range of values for $\rho$ and $p_\tx{rel}$.
Parameter $\rho$ describes how strongly the relevant features are predictive about $y$, so when $\rho$ is close to 1, they all are almost perfect copies of $f$ and therefore easy to distinguish from the noise features.
On the other hand when $\rho$ gets smaller, the predictive power of the relevant features decreases and hence they are more difficult to identify.
It is quite remarkable that above $\rho=0.4$ the reference model approach gives nearly oracle results.


\subsection{Note on the terminology}
\label{sec:selection_terminology}

To avoid confusion, it is useful to distinguish between two different problems both of which could be considered as ``feature selection'':
\begin{enumerate}
	\item Find a \emph{minimal} subset of features that yield a good predictive model for $y$, so that adding more features does not considerably improve predictive accuracy.
	\item Identify \emph{all} features that are predictive about (that is, statistically related to) the target variable~$y$.
\end{enumerate}
In the remainder of this paper, we shall focus on the first problem.
The latter problem---which is often a considerably more difficult one---is usually referred to as \emph{multiple hypothesis testing}, and different means are more suitable for solving that.
Still, as the previous example illustrates (Sec.~\ref{sec:intro_example}), we expect the reference model approach to be beneficial also there.
We shall touch upon this issue in the final discussion (Sec.~\ref{sec:multiple_hypothesis}).

\section{Traditional Bayesian approaches}
\label{sec:traditional_approaches}

This section briefly reviews some of the most common Bayesian approaches for inference with large number of features and highlights their main difficulties.

\subsection{Sparsifying priors}

Consider the standard Gaussian linear regression model 
\begin{align}
\begin{split}
	y_i &= \vs \beta^\tp \vc x_i + \varepsilon_i, \quad \varepsilon_i \sim \Normal{0,\sigma^2}, \quad i=1,\dots,n \,, 
\end{split}
\label{eq:lgm}
\end{align}
where $\vc x$ is the $p$-dimensional vector of features, $\vs \beta$ contains the corresponding regression coefficients and $\sigma^2$ is the noise variance.
A very popular Bayesian approach for assessing the relevances of the different features is to assign a sparsifying prior on each $\beta_j$, and then perform the relevance assessment based on the marginal distributions for each $\beta_j$ (see Sec.~\ref{sec:marg_relevances}).

A popular prior choice is the \emph{spike-and-slab}, which is often written as a mixture of two Gaussians 
\begin{align}
\begin{split}
	\beta_j\given\lambda_j,c,\,\varepsilon &\sim \lambda_j\,\Normal{0,c^2} + (1-\lambda_j)\, \Normal{0,\varepsilon^2}, \\
	\lambda_j \given \pi &\sim \Ber{\pi}, \qquad j=1,\dots,p,
\end{split}
\label{eq:ss_prior}
\end{align}
where $\varepsilon \ll c$ and the indicator variable $\lambda_j\in \{0,1\}$ denotes whether the coefficient $\beta_j$ is close to zero (comes from the ``spike'', $\lambda_j=0$) or nonzero (comes from the ``slab'', $\lambda_j=1$).
The width of the spike $\varepsilon$ is either taken to be exactly zero or set to a small positive value~\citep{george1993,ishwaran2005}.
The prior inclusion probability $\pi$ is either fixed (typically to $\pi=0.5$) or given a hyperprior such as $\pi \sim \Unif{0,1}$ \citep{ishwaran2005}.
In some variants, the Gaussians are replaced by more heavy-tailed distributions, such as Laplacians~\citep{johnstone2004}.

A popular alternative to the spike-and-slab is to formulate the prior for $\beta_j$s as a continuous mixture of Gaussians.
This approach can be computationally more appealing and can avoid some issues that are due to sensitivity to the choices for $\varepsilon$, $c$ and $\pi$.
Several such priors have been proposed \citep[e.g.][]{carvalho2010,armagan2011,bhattacharya2015,bhadra2017}, but the most popular one is probably the \emph{horseshoe}
\begin{align}
\begin{split}
	\beta_j \given \lambda_j,\tau &\sim \Normal{0, \tau^2 \lambda_j^2}, \\
	\lambda_j &\sim \halfCauchy{0,1} \,, \quad j=1,\dots,p,
\end{split}
\label{eq:hs_prior}
\end{align}
which has been shown to possess several attractive properties and has enjoyed a great empirical success \citep{carvalho2009, polson2011, vanderpas2014}.
The intuition is that the global scale $\tau$ drives all the coefficients toward zero, while the thick Cauchy-tails for the local scales $\lambda_j$ allow some of the coefficients to escape the shrinkage.
\cite{piironen2017c} proposed an extension to the formulation~\eqref{eq:hs_prior}, called the \emph{regularized horseshoe}
\begin{align}
\begin{split}
	\beta_j \given \lambda_j,\tau,c &\sim \Normal{0, \tau^2 \ti \lambda_j^2},  \quad \ti \lambda_j^2 =  \frac{c^2 \lambda_j^2}{c^2 + \tau^2 \lambda_j^2}, \\
	\lambda_j &\sim \halfCauchy{0,1} \,, \quad j=1,\dots,p, \\
	c^2 &\sim \InvChi{\nu, s^2},
\end{split}
\label{eq:rhs_prior}
\end{align}
which introduces an additional regularization parameter $c$ that brings the characteristics of the horseshoe even closer to those of the spike-and-slab~\eqref{eq:ss_prior}.
The idea is that unlike in the original horseshoe where the largest coefficients are only very weakly penalized (horseshoe has Cauchy-tails), here they face a regularization equivalent to a Student-$t$ slab with scale $s$ and $\nu$ degrees of freedom.
For a fixed but finite slab width $c=s$ (obtained by letting $\nu \rightarrow \infty$), the prior is operationally similar to the spike-and-slab~\eqref{eq:ss_prior} with the same $c$, whereas the original horseshoe~\eqref{eq:hs_prior} (obtained by letting also $s \rightarrow \infty$) resembles the spike-and-slab with infinite slab width $c \rightarrow \infty$ \citep[see][for the derivations, more detailed discussion and illustrations]{piironen2017c}.
This additional regularization is useful if the parameters are weakly identified (e.g. coefficients in separable logistic regression) and often robustifies and speeds up the Markov chain Monte Carlo (MCMC) posterior inference.

It is possible to place a prior for the global parameter $\tau$ based on the sparsity assumptions analogous to the prior for $\pi$ in spike-and-slab~\eqref{eq:ss_prior}.
Under certain assumptions, \cite{piironen2017b,piironen2017c} showed that to concentrate prior mass onto solutions where $p_0$ coefficients are far from zero, most of the prior mass for $\tau$ should be concentrated near the reference value
\begin{align}
	\tau_0 = \frac{p_0}{p-p_0} \frac{\sigma}{\sqrt{n}}.
\label{eq:tau0}
\end{align}
A recommended weakly informative prior is then $\tau \given \sigma \sim \halfCauchy{0,\tau_0^2}$, which we shall also use throughout this paper unless otherwise stated.

\subsection{Bayes factors and marginal posterior relevance assessment}
\label{sec:marg_relevances}

It should be made explicit that neither the spike-and-slab~\eqref{eq:ss_prior} nor the (regularized) horseshoe~\eqref{eq:rhs_prior} performs actual feature \emph{selection} in the sense that some of the variables would have exactly zero coefficient with probability one, which is true for many of the non-Bayesian penalized estimators (see Sec.~\ref{sec:search_heuristics}).
Although often overlooked, the actual selection problem can remain highly non-trivial even after successfully fitting the model with a sparsifying prior. 

In the spike-and-slab literature, the actual selection is most often carried out either by selecting the most probable feature combination (that is, using Bayes factors) or by selecting those features with posterior inclusion probability above some threshold, typically 0.5, although several thresholding rules have been proposed~\citep{ishwaran2005,narisetty2014}.
The selection based on posterior inclusion probabilities is known to yield a submodel which minimizes the expected squared predictive error under some fairly strict and unrealistic assumptions, most notably that the model and prior are correct, and the features are orthogonal \citep{barbieri2004}.
Analogous decision rule based on the posterior estimates for the so called shrinkage factors could also be devised for the horseshoe~\citep{carvalho2010}.
This is essentially equivalent to simply investigating the marginal posteriors of the regression coefficients, and then choosing those features with the coefficient posterior mass significantly away from zero with some pre-defined credible level.

Unfortunately both the Bayes factors and the marginal relevance assessment have difficulties that make them unsatisfactory in our opinion.
First of all, the posterior inference via MCMC for multimodal posterior resulting from one of the sparsifying priors can be a great challenge for high-dimensional feature spaces.
Even when the posterior inference would not be a problem, the prior sensitivity of the Bayes factors has been long known \cite[see, e.g.,][]{jeffreys1961book, kass1995} and the approach does not lend itself to the continuous shrinkage priors.
In addition, for large number of features $p$ the Bayes factors typically have high Monte Carlo errors due to the fact that only a vanishingly small proportion of the $2^p$ models is visited during MCMC, and almost all models are not visited at all.
The relevance assessment based on the marginal posteriors on the other hand can produce unintuitive results in the case of correlating features, since it can be that the marginals of two or more coefficients overlap with zero but the joint distribution is clearly distinguished from zero (see Sec.~\ref{sec:toy_example}).
Another major issue is that neither of these approaches provides a satisfactory answer to how to perform \emph{post-selection inference} for the selected model, in particular, how to make inference and predictions after the selection, conditional on all the information available. 
This makes it also problematic to perform tradeoff analysis between the number of included variables and the model accuracy (that is, how much predictive accuracy would be gained or lost if one or more features were included or excluded).
For an example of how the projective approach can improve predictions using the selected model even when marginal posterior probabilities are used for selecting the features, see Figure 6 in \cite{piironen2017a}.

 \subsection{An illustrative example}
\label{sec:toy_example}

We illustrate the difficulties with the marginal relevance assessment discussed in Section~\ref{sec:marg_relevances} with similar data as in the introductory example, see Equation~\eqref{eq:toy_data}.
We generated one data realization with $n=50$ observations for three different number of features, $p=4$, $p=10$ and $p=50$, each using $\rho = 0.8$ and $p_\tx{rel} = \frac{p}{2}$, so in each case the first half of the features were truly relevant.
For illustration purposes, we did this by first generating the data for $p=4$ and then adding the right number of relevant and irrelevant features for cases $p=10$ and $p=50$.
This way, the realized values for the first two relevant features $x_1$ and $x_2$ and the target variable $y$ did not vary between the three data sets, which lets us illustrate how the total number of features $p$ affects the relevance assessment of the two features.

\begin{figure}[t]
	\centering
	\minput[pdf]{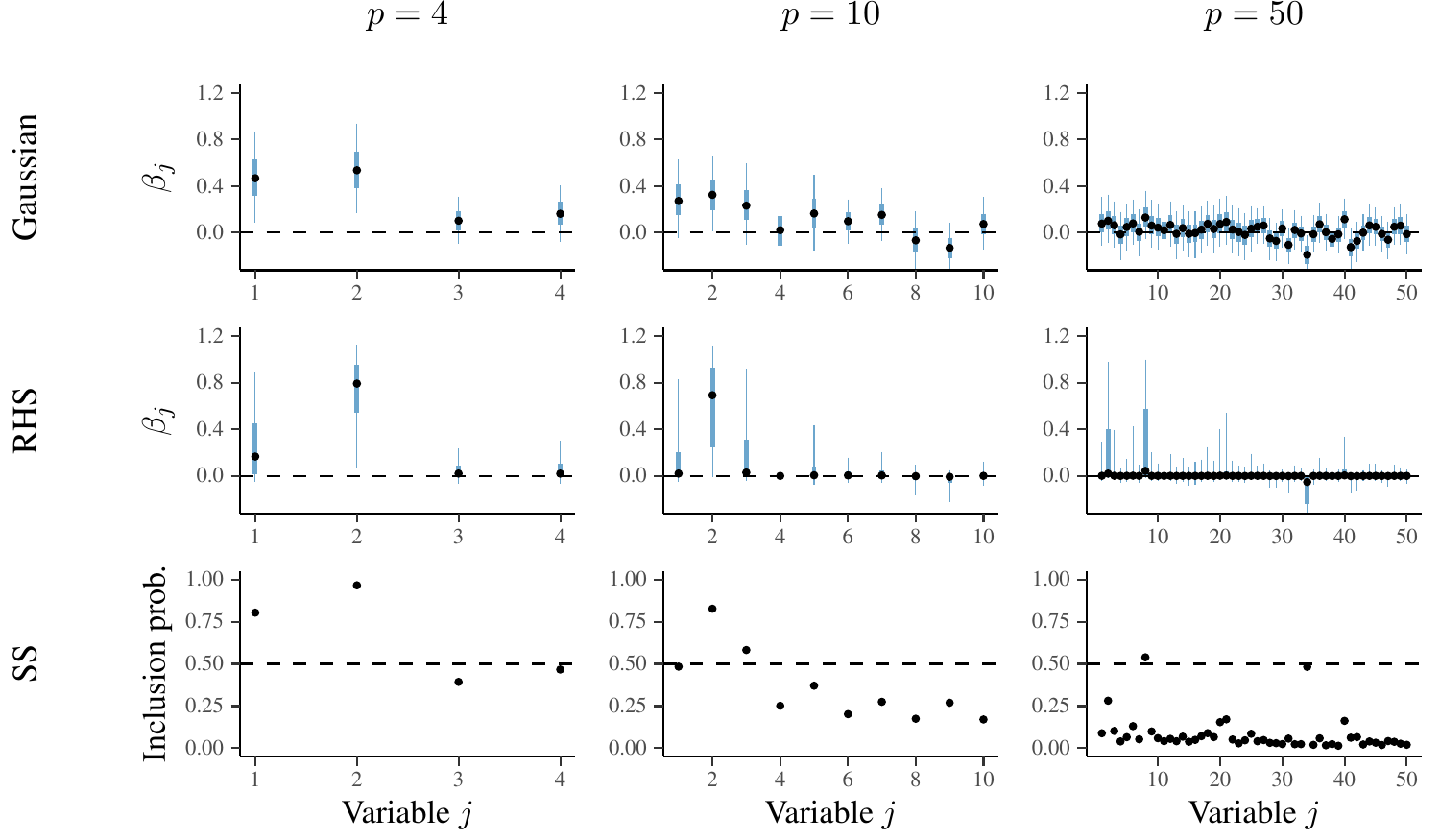}
	\captionspace
	\caption{ {\it Simulated example:} The rows denote the results for the three different priors, Gaussian, regularized horseshoe (RHS) and spike-and-slab (SS), and the columns show the results for the three different number of features $p$. For the Gaussian and RHS priors the graphs show the posterior median (dots) with 50\% and 90\% credible intervals (thick and slim lines, respectively) for the regression coefficients $\beta_j$. For SS prior, the graphs show the posterior inclusion probabilities for each variable. As the dimensionality $p$ increases, all the marginals start to overlap with zero, and the SS posterior inclusion probabilities get smaller.}
	\label{fig:toy_marginals}
\end{figure}

A Bayesian linear regression model was fitted to these data with three different priors on the regression coefficients: 
\begin{itemize}
	\item Gaussian $\beta_j \given \tau \sim \Normal{0,\tau^2}$ with $\tau \sim \halfCauchy{0,1}$
	\item Regularized horseshoe (RHS) with $p_0=1$, $\nu=4$, $s^2=1$ (See Eq.~\eqref{eq:rhs_prior} and \eqref{eq:tau0})
	\item Spike-and-slab (SS)\footnote{For inference, we used the R-package {\tt spikeslab} \citep{ishwaran2010}.} with $\pi \sim \Unif{0,1}$
\end{itemize}
Figure~\ref{fig:toy_marginals} visualizes the posterior median and credible intervals for the regression coefficients under Gaussian and RHS priors, along with the marginal posterior inclusion probabilities for the different features obtained from the SS-posterior.
With only $p=4$ features and Gaussian prior, both $x_1$ and $x_2$ are detected to be relevant as the marginal posteriors of $\beta_1$ and $\beta_2$ are distinguished from zero.
As the number of features grows, the marginals become more overlapping with zero and with $p=50$ the marginals of all the relevant features are substantially overlapping with zero.
The same applies also for the RHS prior, in fact it appears that the marginals start to concentrate around zero faster than for Gaussian prior.
Also for the SS prior, the marginal inclusion probabilities generally decrease for all the relevant features as the dimensionality grows, and for $p=50$ only one of them just barely has probability over 0.5.
Notice how the marginals of the coefficients for the relevant variables are not substantially different from those of the irrelevant ones when $p=50$ regardless of the prior.

\begin{figure}
	\centering
	\minput[pdf]{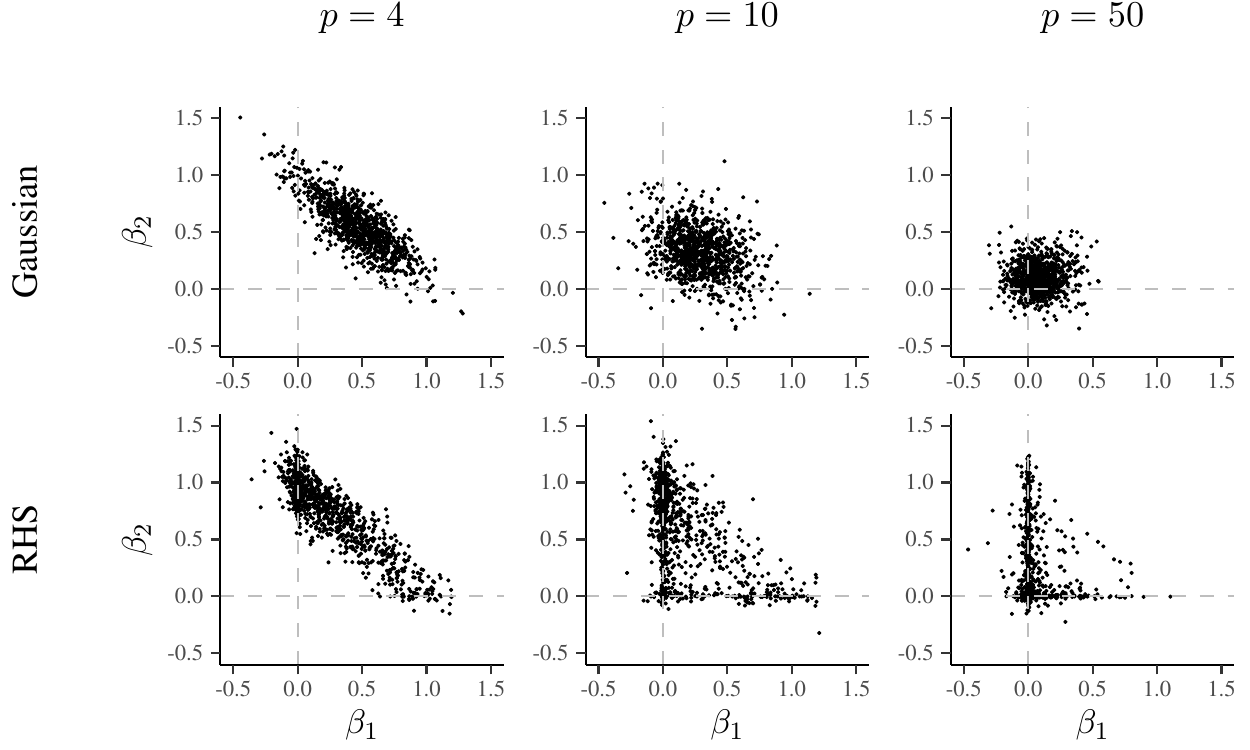}
	\caption{ {\it Simulated example:} Posterior draws for $\beta_1$ and $\beta_2$ with Gaussian and regularized horseshoe (RHS) priors (top and bottom row, respectively) when the total number of features $p$ varies. In each graph, the observed data for $x_1,x_2$ and $y$ are exactly the same, only the prior and the total number of features $p$ varies. Notice how the marginal posteriors are always more overlapping with zero than the joint posterior. As the dimensionality increases (in particular, when the number of features correlating with $x_1$ and $x_2$ increases), the joint posterior becomes more closer to the product of the two marginals and more overlapping with zero.} 
	\label{fig:toy_b1b2}
\end{figure}
\begin{figure}
	\centering
	\minput[pdf]{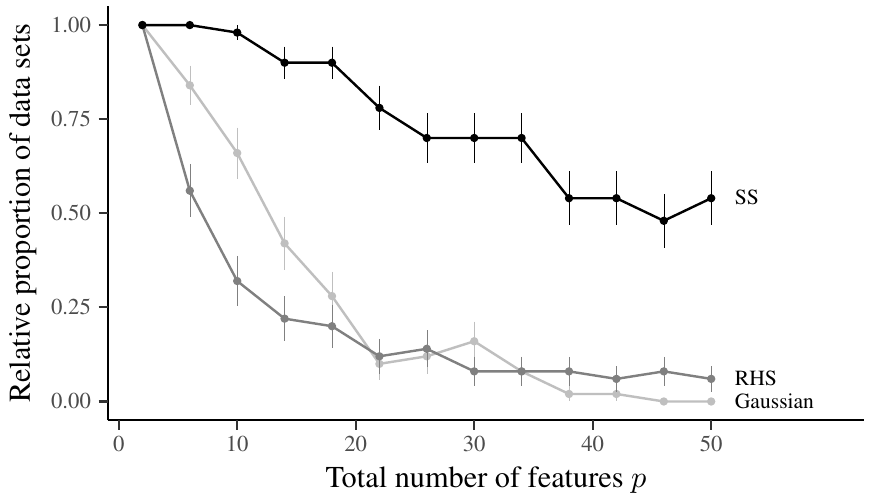}
	\caption{ {\it Simulated example:} Relative proportion of data sets where at least one feature is found to be significant. With SS prior feature is considered significant if its posterior probability exceeds 0.5 and with Gaussian and RHS priors if its coefficient is either positive or negative with posterior probability 0.95 or more. The results are computed from 50 randomly generated data sets generated according to \eqref{eq:toy_data} with $n=50$, $\rho = 0.8$ and $p_\tx{rel} = \frac{p}{2}$. Vertical bars denote one standard error intervals} 
	\label{fig:toy_selection_rep}
\end{figure}

The reason for this behaviour is quite simple: as the number of features carrying similar information grows, the coefficients of most of the relevant features could be set to zero as long as one (or a few) of them obtain nonzero coefficient.
In other words, none of the features is so precious that it could not be removed, and therefore the marginals of all the features become more overlapping with zero.

Figure~\ref{fig:toy_b1b2} further illustrates what happens to the posterior of $\beta_1$ and $\beta_2$ when the dimensionality changes.
For $p=4$ where $x_1$ and $x_2$ are the only relevant features, the posterior dependency between their coefficients is very strong; if one of the coefficients is set to zero, then the other one must be large.
As the number of features $p$ grows, the posterior dependency between $\beta_1$ and $\beta_2$ becomes weaker; when there are many features that carry similar information as $x_1$ and $x_2$, both coefficients could be set to zero because there are many substitutes.
The results for $p=50$ really summarize why the marginals and the pairwise posterior plots can be very challenging to interpret and even misleading: $x_1$ and $x_2$ have correlation of $\rho=0.8$ and their correlation with $y$ both exceed 0.6,\footnote{The correlation between each relevant $x_j$ and $y$ is $\sqrt{\frac{\rho}{2}} \approx 0.63$} yet there is no apparent posterior dependency and both marginals clearly overlap zero!

Figure~\ref{fig:toy_selection_rep} simply confirms that these observations are not due to cherry-picking a specific data set.
For each of the three priors the relative propotion of data sets where at least one feature is found to be significant goes down when $p$ increases.
With Gaussian and RHS priors this probability is already fairly close to zero with $p=50$, and even with SS we fail to find any relevant features in about half of the data sets.
The exact proportions are naturally dependent on the selected thresholding rules (posterior probability of 0.5 in SS and credible level 0.95 for Gaussian and RHS) but these do not affect the main conclusions.

\subsection{Why not to use cross-validation for selecting the feature combination?}

Cross-validation (CV) and information criteria (IC) are widely used generic methods for estimating predictive performance of essentially any learning algorithm. 
One might be wondering why not to use them also for feature selection? 
While it is certainly true that for example cross-validation can be a robust and convenient method for comparing a few competing models, in feature selection the number of model comparisons becomes quickly impractically large even for a relatively small number of candidate features.
The computational burden of fitting a large number of models becomes an obvious problem especially if Bayesian approach with MCMC is used for inference.
Another complication is that the prior needs to be specified separately for each model.
On the other hand, in projective approach (Sec.~\ref{sec:projection}) the prior must be specified only for the reference model and projecting the reference model posterior onto the submodels is usually hugely faster than performing MCMC for the submodels.

Another problem that is not always so well understood is that when many models are compared using cross-validation, the selection process is liable to overfitting which can lead to selection of non-optimal model due to relatively high variance in the cross-validation estimates.
We have discussed this in detail in our earlier work~\citep{piironen2017a} where we also show that the projective approach is considerably more resilient to this phenomenon.
The selection induced bias has also been discussed by other authors, see for example~\cite{ambroise2002}, \cite{reunanen2003} and \cite{cawley2010}.

\section{Predictive projection}
\label{sec:projection}

This section discusses the projective approach in detail.
We start by describing the projective idea in general, and then discuss the exponential family models and GLMs as special cases.

\subsection{Remarks on notation}

We shall denote the training data by $\data$.
The `tilde' notation is used to denote future measurements, for example symbol $\ti y$ denotes unseen measurement for $y$.
To simplify notation, we use $\ti y_i$ to denote a new observation at the $i$th observed feature values~$\vc x_i$, which allows us to drop the conditioning on~$\vc x_i$ from the conditional distributions.
Notice though that $\ti y_i$ is in general different from the observed $y_i$.

\subsection{General idea}
\label{sec:projection_idea}

In generic terms, {\it posterior projection} refers to a procedure of replacing the posterior distribution $p(\vs \theta_* \given \data)$ of the reference model with a simpler distribution $q_\perp(\vs \theta)$ that is restricted in some way.
For example, in feature selection context for GLMs, this would mean constraining some of the regression coefficients to be exactly zero.
In general the domain of the projected parameters $\vs \theta \in \vs \Theta$ can and typically will be different from the domain of the reference model parameters $\vs \theta_* \in \vs \Theta_*$.
For this reason, it is not meaningful to define the projection directly via the discrepancy between $p(\vs \theta_* \given \data)$ and $q_\perp(\vs \theta)$. 
Instead, a natural approach would be to define it via the discrepancy between the induced {\it predictive} distributions
\begin{align}
	\KL{p(\ti y \given \data)}{q(\ti y)}
	&= \phantom{-} \mean[\ti y]{ \log p(\ti y \given \data) - \log q(\ti y)}  \nonumber \\
	&= - \mean[\ti y]{ \log q(\ti y)} + \tx{const.} \nonumber \\
	&= - \mean[\ti y]{ \log \mean[\vs \theta]{p(\ti y \given \vs \theta) } }  + \tx{const.} \nonumber \\
	&= - \mean[\vs \theta_*]{ \mean[\ti y \given \vs \theta_*]{ \log \mean[\vs \theta]{p(\ti y \given \vs \theta) } } } + \tx{const.} 
\label{eq:kl_distr}
\end{align}
Here $\mean[\vs \theta_*]{\cdot}$, $\mean[\ti y \given \vs \theta_*]{\cdot}$ and $\mean[\vs \theta]{\cdot}$ denote expectations over $p(\vs \theta_* \given \data)$, $p(\ti y \given \vs \theta_*)$ and $q_\perp(\vs \theta)$, respectively.
Optimal projection of posterior $p(\vs \theta_* \given \data)$ from parameter space $\vs \Theta_*$ to $\vs \Theta$ in terms of minimal predictive loss would then be the distribution $q_\perp(\vs \theta)$ that minimizes functional~\eqref{eq:kl_distr}.
In practice minimizing this is difficult even for relatively simple models and projected posterior $q_\perp(\vs \theta)$ due to the many expectations, but expression~\eqref{eq:kl_distr} serves as the ideal when re-formulating the projection in a more tractable way.
Below we define three different projections.

\subsection{Practical projection techniques}
\label{sec:projection_methods}

\noindent\textbf{Draw-by-draw\quad}
Instead of trying to minimize the functional \eqref{eq:kl_distr} assuming some parametric form for $q_\perp(\vs \theta)$, we can obtain an easier optimization problem by formulating the projection as a pointwise mapping from a given $\vs \theta_* \in \vs \Theta_*$ to  $\vs \theta_\perp \in \vs \Theta$ as 
\begin{align}
	\vs \theta_\perp
	&= \arg \min_{\vs \theta \in \,\vc \Theta}
	 \KL{p(\ti y \given \vs \theta_*)} { p(\ti y \given \vs \theta) } \nonumber \\
	&= \arg \max_{\vs \theta \in \,\vc \Theta}
	\mean[\ti y \given \vs \theta_*]{ \log p(\ti y \given \vs \theta) }.
\label{eq:proj_pointwise}
\end{align}
For models where the predictions are conditioned on some set of observed predictors $\vc {\ti x}$, one takes the average of~\eqref{eq:proj_pointwise} over the distribution of the predictors.
As the distribution of the future predictors $p(\vc{\ti x})$ is typically not available, the expectations over this are most conveniently approximated by a sample mean over the observed $\{\vc x_i\}_{i=1}^n$.
This results in a projection equation
\begin{align}
	\vs \theta_\perp
	&= \arg \max_{\vs \theta \in \,\vc \Theta}
	 \frac{1}{n}\sum_{i=1}^n \mean[\ti y_i \given  \vs \theta_*]{ \log p(\ti y_i \given \vs \theta) },
\label{eq:proj_pointwise_x}
\end{align}
which is the original formulation of \citep{goutis1998,dupuis2003} (they used minimization of KL-divergence in their formulation, but this is equivalent to maximizing the expected likelihood in Eq.~\eqref{eq:proj_pointwise_x}).
Given draws $\{\vs \theta_*^s\}_{s=1}^S$ from the posterior $p(\vs \theta_* \given \data)$ we can project each of these separately via~\eqref{eq:proj_pointwise_x} to obtain the corresponding draws $\{\vs \theta_\perp^s\}_{s=1}^S$ in the projection space~$\vs \Theta$.
These can be thought of as draws from a projected posterior distribution $q_\perp(\vs \theta)$ (although this may not be available analytically), and hence they are used exactly as we would use posterior draws for that particular submodel.
The appealing property of the draw-by-draw projection is that it is computationally feasible for many commonly used models such as the GLMs because the optimization problem will have the same form as the problem of finding the maximum likelihood parameter values (see Sec.~\ref{sec:glm}).
The introduced projection error or loss is then defined as the average loss over the draws
\begin{align}
	\delta_{\vc \Theta} = \frac{1}{S} \sum_{s=1}^S 
							\KL{p(\ti y \given \vs \theta_*^s)} { p(\ti y \given \vs \theta_\perp^s) }.
\end{align} \\

\noindent\textbf{Single point (one cluster)\quad} 
Draw-by-draw projection (above) maps each parameter value $\vs \theta_*$ into a corresponding value $\vs \theta_\perp$ in the projection space.
The single point projection (which is a special case of the clustered projection that we will introduce in a moment) instead maps the whole posterior $p(\vs \theta_* \given \data)$ into a single value $\vs \theta_\perp$.
This can be obtained from~\eqref{eq:kl_distr} by assuming $q_\perp(\vs \theta)$ is a point mass at~$\vs \theta \in \vs \Theta$, taking expectation over the predictors $\vc{\ti x}$ and then optimizing the expression with respect to~$\vs \theta$
\begin{align}
	\vs \theta_\perp
	&= \arg \max_{\vs \theta \in \,\vc \Theta}
	 \frac{1}{n}\sum_{i=1}^n \mean[\ti y_i]{ \log p(\ti y_i \given \vs \theta) }.
\label{eq:proj_point}
\end{align}
This is the formulation of~\cite{tran2012}.
Notice that~\eqref{eq:proj_point} is otherwise same as \eqref{eq:proj_pointwise_x} except that here the expectation is computed over the posterior predictive distribution of the reference model, that is,  $\mean[\ti y_i]{\cdot} = \mean[\vs \theta_*]{\mean[\ti y_i \given \vs \theta_*]{\cdot}}$, where $\mean[\vs \theta_*]{\cdot}$ denotes expectation over $p(\vs \theta_* \given \data)$.
In practice the expectation $\mean[\ti y_i]{\cdot}$ is approximated using the posterior draws.
Equation~\eqref{eq:proj_point} can be used to compute optimal point estimates in the projection space.
Also, when $\vs \Theta = \vs \Theta_*$ this computes the optimal predictive point estimates in the original parameter space~\citep[for a related approach, see][]{bernardo2003}.
It is worth noticing that in general the result is often different from the usual point estimates, such as the posterior mean or median.

The benefit of the single point projection over the draw-by-draw is that it is much lighter computationally. 
For instance, for GLMs (Sec.~\ref{sec:glm}), solving \eqref{eq:proj_point} has the same computational complexity as solving \eqref{eq:proj_pointwise_x}, and since the latter must be solved separately for each of the~$S$ posterior draws, single point projection essentially reduces the computations by a factor of~$S$.
Another benefit of formulation~\eqref{eq:proj_point} is that it allows convenient search techniques, such as the Lasso type $L_1$-penalty, to be used for finding good submodels~\citep{tran2012}.
We will discuss this more closely in Section~\ref{sec:search_heuristics}. 
The drawback is that it can be somewhat less accurate than the one-to-one projection, meaning that the predictive accuracy of the submodel can be compromised.
To address this point, we shall introduce the clustered projection below. \\

\noindent\textbf{Clustered \quad} The clustered projection is our novel approach that can be thought of as a unification of the draw-by-draw and single point projections. 
In this approach one clusters the posterior draws~$\{\vs \theta_*^s\}_{s=1}^S$ of the reference model into $C$ clusters $\{\vs \theta_*^s : s \in I_c\}, \,\, c=1,\dots,C$, and then performs a single point projection within each cluster.
Here $I_1,\dots, I_C$ denote the index sets that indicate which draw belongs to which cluster (we discuss in a moment how to come up with such a division).
The projection for the $c$th cluster then becomes
\begin{align}
	\vs \theta_\perp
	&= \arg \max_{\vs \theta \in \,\vc \Theta}
	 \frac{1}{n}\sum_{i=1}^n \mean[\ti y_i \given I_c]{ \log p(\ti y_i \given \vs \theta) },
\label{eq:proj_clustered}
\end{align}
where $\mean[\ti y_i | I_c]{\cdot}$ denotes the predictive distribution of the reference model computed over the posterior draws in that cluster $I_c$.
In other words, $\mean[\ti y_i | I_c]{h(\ti y_i)} = \frac{1}{|I_c|}\sum_{s \in I_c} \mean[\ti y_i\given \vs \theta_*^s]{h(\ti y_i)}$ for any function~$h(\ti y_i)$.\footnote{Here we are slightly abusing the notation by using the symbol $\mean{\cdot}$ to denote sample mean computed over a finite number of posterior draws, but we do this to simplify the notation.}
Solving~\eqref{eq:proj_clustered} for each or the $C$ clusters yields a set of projected parameters $\{\vs \theta_\perp^c\}_{c=1}^C$.
Each of these is given a weight $\omega_c$ proportional to the number of draws in that cluster, $\omega_c = \frac{|I_c|}{S}$, and these weights are taken into account when computing expectations over the projected posterior. For example, the projected predictive density at future $\ti y$ is then given by
\begin{align}
	q(\ti y) = \sum_{c=1}^C \omega_c \, p(\ti y \given \vs \theta_\perp^c).
\label{eq:pred_clustered}
\end{align}
More generally, the expectation of an arbitrary function $h(\vs \theta_\perp)$ over the projected posterior is calculated as $\sum_{c=1}^C \omega_c h(\vs \theta_\perp^c)$.

A simple but generic and effective approach is to cluster the draws~$\{\vs \theta_*^s\}_{s=1}^S$ based on the expected values they impose for $y$ in the unconstrained (latent) space.
That is, if $\vc f_s = g(\mean{\vc {\ti y} \given \vs \theta_*^s})$, where $\vc {\ti y}=(\ti y_1,\dots,\ti y_n)$ and~$g(\cdot)$ denotes the link function, we would cluster the vectors $\{\vc f_s\}_{s=1}^S$.
This approach is convenient since it makes the clustering independent of the dimensionality of the parameter space of the reference model, and since in practice for projection we need only the vectors~$\vc f_s$ (see Sec.~\ref{sec:exponential_family} and~\ref{sec:glm}), we can perform the clustering with access only to the predictions of the reference model (without access to the actual parameter values).
As a clustering algorithm, we use $k$-means.
An alternative approach would be to minimize the locations of the projected parameters $\{\vs \theta_\perp^c\}_{c=1}^C$ jointly using for example the method of \cite{snelson2005}, but this is computationally much more expensive.

Both the draw-by-draw~\eqref{eq:proj_pointwise_x} and the single point projection~\eqref{eq:proj_point} are obtained as special cases of the clustered projection~\eqref{eq:proj_clustered}.
The draw-by-draw approach is obtained by setting the number of clusters $C$ equal to the number of posterior draws $C=S$ and assigning each posterior draw into its own cluster.
The single point projection is obtained by setting $C=1$ and assigning all draws into the same cluster.
The benefit of the clustered projection is that it improves the accuracy compared to the single point (one cluster) projection already with a small number of clusters, and thereby gives a good tradeoff between speed and accuracy.
We will illustrate this with an example in Sec.~\ref{sec:clusterdemo}.

\subsection{Exponential family models}
\label{sec:exponential_family}

Assuming the observation model for $y_i$ belongs to the exponential family with canonical parameter~$\xi_i$ and dispersion $\phi$, the log-likelihood has the form \citep[][ch.~2]{mccullagh1989book}
\begin{align}
	\mc L_i = \log p(y_i \given \xi_i) = \frac{y_i \xi_i - B(\xi_i)}{A(\phi)} + H(y_i,\phi),
\label{eq:expfam_ll}
\end{align}
for some specific functions $A(\cdot)$, $B(\cdot)$ and $H(\cdot)$.
Here the natural parameter is a function of the model parameters, $\xi_i = \xi_i(\vs \theta)$.
The maximum likelihood solution for the parameters $\vs \theta$ reduces to 
\begin{align}
	\vs \theta_\tx{ML} 
	= \arg \max_{\vs \theta \in \,\vc \Theta} \sum_{i=1}^n \big(y_i \xi_i(\vs \theta) - B(\xi_i(\vs \theta)) \big),
\label{eq:exponential_ml}
\end{align}
which does not depend on the value for the dispersion $\phi$ (function $A(\phi)$ is assumed to be strictly positive).
Let $\ti y_i$ denote a new measurement at the $i$th observed feature values $\vc x_i$.
Now, if we denote the expected value of $\ti y_i$ over some reference distribution as $\mu_i^* = \mean{\ti y_i}$, we can write the draw-by-draw, single point and clustered projections (Eq.~\eqref{eq:proj_pointwise_x}, \eqref{eq:proj_point} and~\eqref{eq:proj_clustered}) all as 
\begin{align}
	\vs \theta_\perp
	&= \arg \max_{\vs \theta \in \,\vc \Theta}
	 \sum_{i=1}^n \big(\mu_i^* \xi_i(\vs \theta) - B(\xi_i(\vs \theta)) \big).
\label{eq:exponential_proj}
\end{align}
Thus when the observation model of the submodel is in the exponential family, the projection of the model parameters $\vs \theta$ is equivalent to finding the maximum likelihood solution with the observed targets $\vc y=(y_1,\dots,y_n)$ replaced by their expected values $\vs \mu_* = (\mu_1^*,\dots, \mu_n^*)$ as predicted by the reference model.
Thus the projection can be considered as ``fitting to the fit'' of the reference model.
As discussed in Section~\ref{sec:projection_methods}, in draw-by-draw projection these fitted values $\mu_i^*$ are computed separately for each posterior draw in the reference model, in clustered projection separately for each cluster, and ultimately in the one cluster (single point) projection over the whole posterior with the parameters $\vs \theta_*$ integrated out.
Notice also that the projection of the parameters $\vs \theta$ does not depend on the value for the dispersion parameter $\phi$.

It is worth emphasizing that this result assumes only that the observation model of the reduced model belongs to exponential family.
In particular, we are not making any assumptions about the observation model of the reference model (which need not belong to the exponential family) or about the functional form of $\xi(\vs \theta)$ or about how the reference fit $\vs \mu_*$ is formed.
In principle this means that the projection could be applied to a wide class of learning algorithms simply by plugging in the fit of the reference model in place of the observed targets $y_i$ in maximum likelihood estimation. 
In practice, though, this does not work for nonparametric models such as Gaussian processes where the parameters are the values $\xi_i$ themselves without further assumptions.

After computing the projected values for the model parameters $\vs \theta$ (Eq.~\eqref{eq:exponential_proj}), the dispersion $\phi$ is computed from 
\begin{align}
	\phi_\perp
	&= \arg \max_\phi
	 \sum_{i=1}^n \left( \frac{r_i(\vs \theta_\perp)}{A(\phi)} + \mean[\ti y_i]{H(\ti y_i,\phi)} \right),
\label{eq:exponential_proj_disp}
\end{align}
where $r_i(\vs \theta_\perp) = \mu_i^* \xi_i(\vs \theta_\perp) - B(\xi_i(\vs \theta_\perp))$ does not depend on $\phi$.
Again, in draw-by-draw and clustered projection, the expectation in Equation~\eqref{eq:exponential_proj_disp} is computed separately for each draw or cluster, and in single point projection by integrating over the whole posterior.

\subsection{Generalized linear models}
\label{sec:glm}

GLMs have their observation model in the exponential family and thus the discussion of Section~\ref{sec:exponential_family} applies.
Let us first consider the projection onto a linear Gaussian model with feature matrix $\vc X$, where the parameters are the regression coefficients $\vs \beta$ and dispersion is the noise variance~$\sigma^2$.
For simplicity, let us now assume also that the reference model is a linear Gaussian model with feature matrix $\vc Z$ and parameters $(\vs \beta_*, \sigma^2_*)$ and that we have drawn a posterior sample $\{ \vs \beta_*^s, \sigma^2_{*,s}\}_{s=1}^S$.
In terms of Equation~\eqref{eq:expfam_ll}, we have 
\begin{align*}
	\xi_i &= \vs \beta^\tp \vc x_i, \quad 
	A(\sigma^2) = \sigma^2, \\
	B(\xi_i) &= \frac{\xi_i^2}{2}, \quad
	H(y_i,\sigma^2) = -\frac{1}{2} \left( \frac{y_i^2}{\sigma^2} + \log 2\pi \sigma^2 \right).
\end{align*}
Consider now the clustered projection with $C$ clusters.
As discussed in Section~\ref{sec:exponential_family}, the projection solution for $\vs \beta$ within each cluster is obtained by plugging in the fit of the reference model in place of~$\vc y$ into the familiar maximum likelihood solution
\begin{align}
	\vs \beta_c =  (\vc X^\tp \vc X)^{-1} \vc X^\tp \vs \mu_*^c,
\label{eq:proj_coeff}
\end{align}
where $\vs \mu_*^c = \frac{1}{|I_c|}\sum_{s\in I_c} \vc Z \vs \beta_*^s$ denotes the prediction within the $c$th cluster.
In the single point projection ($C=1$) this reduces to $\vs \mu_* = \frac{1}{S}\sum_{s=1}^S \vc Z \vs \beta_*^s$, whereas in the draw-by-draw ($C=S$) we have $\vs \mu^s_* = \vc Z \vs \beta_*^s$.

After plugging~\eqref{eq:proj_coeff} into~\eqref{eq:exponential_proj_disp},
it is straightforward to show that the projection of the noise variance becomes
\begin{align}
	\sigma_c^2
	= \frac{1}{n} \sum_{i=1}^n V_i^c +  \frac{1}{n} ||\vc X  \vs \beta_c - \vs \mu_*^c||^2,
\label{eq:proj_sigma2}
\end{align}
where $V_i^c$ denotes the predictive variance of $\ti y_i$ in the reference model within the $c$th cluster.
This is given by
\begin{align}
	V_i^c = \var{\ti y_i\given I_c}
		&= \mean{\var{ \ti y_i \given \vs \beta_*, \sigma^2_* } \given I_c}
			+ \var{\mean{\ti y_i \given \vs \beta_*, \sigma^2_* } \given I_c} \nonumber \\
		&= \mean{\sigma_*^2 \given I_c} + \var{ \vc z_i^\tp  \vs \beta_*^s \given I_c } \nonumber \\
		&= \frac{1}{|I_c|}\sum_{s\in I_c} \sigma^2_{*,s} + 
			\svar[s \in I_c]{}{ \vc z_i^\tp  \vs \beta_*^s },
\label{eq:predvar}
\end{align} 
where $\svar[s \in I_c]{}{\cdot}$ denotes sample variance over indices $s\in I_c$.
Result~\eqref{eq:proj_sigma2} has a natural interpretation; the projected noise variance is the average predictive variance of the reference model plus the mismatch between the projected and the reference model.
Therefore any systematic variation in the data captured by the reference model but not by the reduced model will be added to the unstructured noise term in the reduced model.
Notice also that the predictive uncertainty of the projected model can never be smaller than in the reference model which shows why the projection provides guard against overfitting in the submodels.

Above we assumed that also the reference model is linear with Gaussian noise.
As already pointed out in Section~\ref{sec:exponential_family}, we emphasize that Equations~\eqref{eq:proj_coeff} and~\eqref{eq:proj_sigma2} hold even without these assumptions.
For instance, $\vs \mu_*^c$ could come from an arbitrary model, such as Gaussian process (GP), neural network or some complex simulation model, and in the projection we investigate how much accuracy is sacrificed by replacing it with a linear model.
Even when the reference model does not account for uncertainty in $\vs \mu_*$, that is, when no clustering can be made, the single point projection is always available for the reference fit $\vs \mu_*$.
Also, the reference model noise could be non-Gaussian---Student-$t$, for instance---but we could still project this model onto a Gaussian noise.

When the observation model of the projected model is non-Gaussian or when the link is non-identity, the maximum likelihood solution is not available analytically, and therefore no closed form solutions for the projected regression coefficients or dispersion parameters exist.
For solving the regression coefficients, the standard approach then is to use iteratively reweighted least squares algorithm (IRLS), where each of the log-likelihood terms $\mc L_i$ is replaced by a pseudo Gaussian observation whose mean and variance are determined either by second order Taylor series expansion to $\mc L_i$ \citep[e.g.][ch.~16.2]{gelman2013book} or by linear approximation to the link function \citep[][ch.~2.5]{mccullagh1989book} at the current iterate (with canonical link functions the two approaches are equivalent).
The process is then iterated until convergence.
Given the solution to the regression coefficients, one can then plug that into Equation~\eqref{eq:exponential_proj_disp} and solve the corresponding value for the dispersion (which might also require an iterative procedure).

\section{Search strategies}
\label{sec:search_heuristics}

Due to the combinatorial explosion, even for relatively small number of features it is impossible to go through all the combinations when finding the optimal reduced model for a given number of features.
Therefore one has to rely on approximate search heuristics for exploring promising submodels.
Probably the simplest alternative is to use a forward stepwise excursion.
This procedure starts from the model with only the intercept term and sequentially adds the feature that decreases the projection error the most.
Forward search can be used together with any of the three projection techniques presented in Section~\ref{sec:projection_methods} and often works well, but it can be computationally expensive for large number of features.

In the case of single point projection~\eqref{eq:proj_point}, a viable alternative is to use either a Lasso-type $L_1$-penalization \citep{tibshirani1996} or the more general elastic net penalty \citep{zou2005} which contains $L_1$-penalty as a special case. 
The single point projection for GLMs with elastic net penalty can be written as
\begin{align}
\min_{\vs \beta}\left\lbrace -\frac{1}{n}\sum_{i=1}^n \mean[\ti y_i]{\mc L_i(\vs \beta,\ti y_i)}
+ \lambda \left( \frac{1}{2}(1-\alpha)||\vs \beta||_2^2  + \alpha ||\vs \beta||_1\right) \right\rbrace.
\label{eq:elnet_projection}
\end{align}
Here the first term is the expectation of the negative of the expected log-likelihood of the submodel with coefficient vector $\vs \beta$ over the predictive distribution of the reference model, and $\alpha$ is the elastic net mixing parameter that bridges the gap between Lasso ($\alpha=1$) and ridge ($\alpha=0$).
Solving this for $\alpha > 0$ over a grid of values for $\lambda$ yields a sequence of models with varying number of regression coefficients different from zero, which can then be used to order the features, for instance by recording the order in which their coefficients break nonzero as $\lambda$ is decreased\footnote{Notice that this is not necessarily the same order in which the coefficients go to zero as the penalty term $\lambda$ is increased. This is because a coefficient that is nonzero can go back to zero as $\lambda$ is reduced, but most of the time the two orderings are the same.}.
It is known that in the case of correlating predictors, Lasso tends to select only one or a few of them discarding the others, while elastic net with $0 < \alpha < 1$ tends to select correlating predictors in groups~\citep{hastie2015book}.
Often $\alpha$ is treated as a higher level parameter and is selected on more subjective grounds.

One of the key advantages of elastic net over the forward stepwise search is that it is computationally very efficient. 
In particular, the coordinate descent algorithm of \cite{friedman2010} that exploits warm starts can often compute the solution path over the entire $\lambda$ grid in comparable time to a single IRLS fit for a fixed variable combination.
We shall not discuss the algorithm but instead refer to the original paper for more information.
However, we do emphasize that unlike in the penalized GLM literature, we use the penalization \emph{only} to find promising submodels, not to regularize their fit after selection.
In other words, after we have solved problem \eqref{eq:elnet_projection} for a grid of values~$\lambda$, we order the features from the most relevant to the least relevant, and find the projected parameter values (or projected posteriors) of the submodels {\it without} any penalization, or using only a small $L_2$-regularization to improve numerical stability.
This is because the projection conditions on the information in the reference model and is therefore much more resilient to overfitting than maximum likelihood estimation for the parameters after selection.
See Section~\ref{sec:clusterdemo} for an illustration of this point, and Section~\ref{sec:benefit_of_reference} for a demonstration of how the predictive accuracy can greatly benefit from not using the penalization for the submodels after selection.

In addition to Lasso and elastic net, there is a wide literature on different penalties for the (generalized) linear models, that are used to induce sparsity in the solution, and therefore could be used as search heuristics to find promising submodels for the projection also.
One such method is the adaptive Lasso \citep{zou2006} which is obtained from \eqref{eq:elnet_projection} by introducing penalty factors $\gamma_j$ that result in different penalization for different variables $\lambda_j = \gamma_j \lambda$,\, $j=1,\dots,p$.
Plugging the local penalties into the regularization term in \eqref{eq:elnet_projection}, the regularizer becomes
\begin{align*}
J(\vs \beta) 
=  \lambda  \sum_{j=1}^p \gamma_j \left( \frac{1}{2}(1-\alpha)  \beta_j^2  + \alpha | \beta_j | \right).
\end{align*}
Using pilot estimates $\vs \beta'$ for the coefficients (that can be the univariate regression coefficients, for example) and setting $\gamma_j = 1/|\beta_j'|^\nu$ for some $\nu > 0$, adaptive Lasso reduces the excessive shrinkage of the relevant coefficients and recovers the true model under more general conditions than does the Lasso.
Adaptive Lasso can also be used to encode preferences for different variables, for instance, due to varying measurement costs.
In the projection context, \cite{tran2012} proposed to set $\vs \beta'$ to the posterior mode of the reference model (assuming it is also a GLM) whereas \cite{hahn2015} proposed to use the posterior mean (the two choices are in general different for GLMs with non-Gaussian priors for the reference model).
Our approach differs from these in that we set $\gamma_j=1$ for each feature in the selection phase but then relax completely $\gamma_j=0$ after the feature selection is done.
We also utilize clustered or draw-by-draw projection after selection when appropriate (see Sec.~\ref{sec:clusterdemo}). 
Another difference to the  approach of Hahn and Carvalho is that they used squared error instead of the KL-divergence to measure the discrepancy to the reference model.
\cite{nott2010} also used $L_1$-penalization but for the draw-by-draw projection.
In this method the different draws can generally project onto different feature combinations even for fixed $\lambda$, and thus this approach does not perform feature selection in the sense we are interested.

Another useful search heuristic is the group Lasso penalty \citep[see, e.g.,][]{hastie2015book} which allows selecting features in groups meaning that all features in the same group are either selected or discarded simultaneously.
Other sparsity enforcing penalties that could be used as search heuristics include the smoothly clipped absolute deviation (SCAD) \citep{fan2001} and the Dantzig selector \citep{candes2007}, but we shall not discuss these further.

\section{Validation and decision rules for model size selection}
\label{sec:validation}

Although we can find the optimal reduced model for a given model complexity by selecting the model with minimal projection loss, making the decision about the appropriate model complexity using the KL-divergences is often difficult.
Firstly, it is not easy to assess how much predictive accuracy is lost for a given amount of projective loss introduced.  
Secondly, since the reference model is never perfect in practice, it is possible to find a submodel with nonzero projection loss but which gives as good predictions as the reference model \citep{piironen2017a}.
Therefore a natural way of deciding the model complexity is to validate the predictive utility of both the reference model and the candidate reduced models on a validation set using a metric that is easy to interpret, and then make the decision based on these validation results.
A generic and useful utility function is the mean log predictive density (MLPD) over the validation points \citep[see, e.g.][]{vehtari2012}, which has the advantage that it measures not only the point predictions but also how well the predictive uncertainties are calibrated.
Various other utility and loss functions could also be used, such as mean squared error (MSE) or classification accuracy in classification problems, which are often easier to interpret.

If plenty of data are available and computation time is an issue, this assessment can be done on hold-out data.
However, when data are scarce, more accurate assessment can be obtained using either leave-one-out (LOO) or $K$-fold cross-validation, which we shall discuss next.

\subsection{$K$-fold cross-validation}
\label{sec:kfold_validation}

In $K$-fold cross-validation both the reference model fitting and the selection is performed $K$ times each time computing the utilities on the corresponding validation set~\citep{peltola2014}. 
This gives us the cross-validated pointwise utilities $u_k^{(i)}$ for a given model complexity $k$ (number of features) at each datapoint $i$. 
For instance, with log predictive density as the utility function, $u_k^{(i)}$ is the log predictive density of the submodel with $k$ features evaluated at the left out $y_i$.
These can then be used to make the final decision about the appropriate level of complexity.
Our approach is to estimate the utility of each model size $k$ relative to the reference model, that is, $\Delta U_k = U_k - U_*$, where~$U_k$ and~$U_*$ denote the true (unknown) utilities for the reduced and the reference model, respectively.
The point estimate and the standard error for the relative utility $\Delta U_k$ in such pairwise comparison are given by
\begin{align}
	\Delta \bar U_k &= \frac{1}{n} \sum_{i=1}^n \left( u_k^{(i)} - u_*^{(i)} \right),
	\label{eq:rel_util_mean} \\
	s_k &= \sqrt{\frac{1}{n} \svar[i=1]{n}{u_k^{(i)} - u_*^{(i)}}},
	\label{eq:rel_util_se}
\end{align}
where $\svar[i=1]{n}{\cdot}$ denotes the sample variance.
Given the point estimate and its standard error it is easy to construct desired confidence intervals for $\Delta U_k$.
A natural choice is then to choose the simplest model that has acceptable difference relative to the reference model with some confidence~\citep{piironen2017a}.

A simple choice is to select the smallest model for which the utility estimate is no more than one standard error away from that of the reference model, that is, the smallest $k$ that satisfies $\Delta \bar U_k + s_k \ge 0$, which means that the submodel is no worse than the reference model with probability approximately $\alpha=0.16$.
This approach has the drawback that such a model is not guaranteed to be found if the submodels all introduce a considerable loss in utility.
Instead one could compare the utilities relative to the best submodel found, that is, in Equation~\eqref{eq:rel_util_mean} replace $u_*^{(i)}$ by $u_{k_\tx{best}}^{(i)}$ where $k_\tx{best} = \arg \max_k \Delta \bar U_k$.
Based on the experiments in Section~\ref{sec:microarray_data} the two choices perform quite similarly, the latter tending to select less parsimonious models but also with slightly better predictive accuracy.
Depending on the application, one might be willing to sacrifice more utility in order to simplify the model ever further, and the decision about the appropriate model size could naturally be made on more subjective grounds also.

\subsection{Leave-one-out cross-validation}
\label{sec:loo_validation}

\subsubsection{Pareto smoothed importance sampling}

The drawback in the $K$-fold cross-validation is that it requires fitting the reference model $K$ times.
Here we propose an alternative approach using approximate leave-one-out (LOO) validation using the Pareto smoothed importance sampling (PSIS) \citep{vehtari2017b}, which avoids the repeated fitting of the reference model. 
In (PS)IS-LOO, the posterior draws can be treated as draws from the LOO posteriors given the importance weights.
The weight for draw $\vs \theta_*^s$ after leaving $i$th observation out, $w_s^{(i)}$, is given by $w_s^{(i)} \propto \frac{1}{p(y_i \given \vs \theta_*^s)}$.
These raw weights are then regularized using Pareto smoothing to stabilize the LOO estimates in case the importance weight distribution has a thick tail \citep[see][for the procedure]{vehtari2017b}.
It is then easy to approximate the desired quantities for the LOO folds using these weights.
For instance, in the clustered projection for the Gaussian linear model we need the predictive means $\vs \mu_*^c$ and variances $(V_1^c,\dots,V_n^c)$ from the reference model for each cluster $c$.
If the reference model is also linear with Gaussian noise, using the notation from Section~\ref{sec:glm}, the predictive means at the observed inputs for the $i$th LOO are given by
\begin{align}
	\vs \mu_*^c = \sum_{s\in I_c} w_s^{(i)} \, \vc Z \vs \beta_*^s,
\end{align}
where the weights are assumed to be normalized $\sum_{s\in I_c} w_s^{(i)} = 1$.
Correspondingly, the predictive variance at point $j$ for the $i$th LOO is given by
\begin{align}
	V_j^{c}
	&= \sum_{s\in I_c} w_s^{(i)} \, \sigma^2_{*,s} + 
			\svar[s \in I_c]{}{ \vc z_j^\tp  \vs \beta_*^s, w_s^{(i)} } %
\label{eq:predvar_loo}
\end{align}
where $\svar[s \in I_c]{}{\, \cdot, v_s}$ denotes the weighted sample variance over indices $s \in I_c$ with weights $v_s$.
Equation~\eqref{eq:predvar_loo} is merely the weighted version of formula~\eqref{eq:predvar}.
The feature selection and the projection onto the submodels at the search path are then carried out for each LOO exactly as in the $K$-fold case.
Exactly the same decision rules as with the $K$-fold validation can be used to decide the appropriate model size, the LOO method simply gives an alternative procedure for computing the pointwise utilities, $u_k^{(i)}$ and $u_*^{(i)}$, for the reduced and reference models, respectively.

PSIS has the benefit that it gives us the Pareto $\hat k$-diagnostics for each LOO describing the accuracy of the importance sampling approximation.
\cite{vehtari2017b} discuss the interpretation of the $\hat k$-values in detail. Based on theoretical and empirical considerations, they conclude that values $\hat k\le 0.7$ indicate reliable approximation.
Larger values indicate that the calculated utilities $u_k^{(i)}$ and $u_*^{(i)}$ for such observation $i$ have high variance and can be biased (optimistic).
In Section~\ref{sec:toy_revisited} we demonstrate empirically that even when a few $\hat k$-values exceed this threshold the relative utility estimate~\eqref{eq:rel_util_mean} can be nearly unbiased since the bias in both $u_k^{(i)}$ and $u_*^{(i)}$ tends to cancel out in the subtraction.

\subsubsection{Subsampling}
\label{sec:subsampling_loo}

Although the (PS)IS-LOO validation avoids the repeated fitting of the reference model, the computation can still get quite involved for large data sets if the selection and projection onto the reduced models is repeated $n$ times.
In such situations it is usually advisable to resort to $K$-fold validation (or hold-out), but an alternative approach would be to compute only a subset of the LOO folds.
Selecting a random subsample of $m < n$ datapoints gives us an unbiased estimate of the submodel utilities but with higher variance than would be obtained by computing all the $n$ LOOs.
This method is analogous to the hold-out method, but with the difference that the full model is learned using all the data.
Since we expect the whole projective idea to be most advantageous when $n$ is small (and hence uncertainties high), we do not focus on large data sets but provide some further ideas of reducing the variance of the subsampling LOO estimate in appendix~\ref{app:subsampling_loo}.

\subsection{Importance of validating the search}
\label{sec:search_validation}

In order to reduce computations, it might be tempting to perform the reference model fitting and feature selection only once using all the available data, and then simply use LOO or $K$-fold CV to estimate the performance of the found submodels.
We strongly advice \emph{not} to employ this strategy, as this is known to produce biased performance estimates, and the bias can be substantial especially for small $n$ and large number of features~\citep[see][for illustrations]{piironen2017a}.
To avoid the selection induced bias, it is important that the same data are never used simultaneously for selection and assessment, meaning that the selection must be performed separately for each of the cross-validation folds regardless of the feature selection method.
Section~\ref{sec:toy_revisited} shows an example of the resulting bias when the selection process is not taken into account in the model assessment.

\section{On the construction of the reference model}
\label{sec:refmodel_construction}

How to construct a good reference model is naturally a central issue in the whole projective approach.
It should be clear that this is essentially an open-ended question with no definite answer; for each problem there are endless choices.
For simple linear and logistic regressions with moderate number (say less than a hundred) features we recommend using all the features with a sparsifying prior, which can work better than a non-sparse prior like Gaussian.
If one is uncertain about the prior, the recommended strategy is to try different choices and compare the resulting fits with cross-validation.

In high-dimensional problems, say with hundreds of features or more, fully Bayesian approach can still provide a good fit but can also prove computationally expensive~\citep{piironen2017c}.  
Using either feature screening, dimension reduction or the combination of the two can be very successful for alleviating the computational burden without sacrificing the predictive accuracy~\citep{neal2006,fan2008,piironen2018}.
In our experience this is true especially for data sets that have plenty of features many of which are correlated with each other and predictive about the target variable.
Microarray data sets (Sec.~\ref{sec:microarray_data}) are typical examples that fall into this category.

For these problems a simple but useful recipe combining feature screening and dimension reduction is known as supervised principal components (SPCs)~\citep{bair2006}, which works as follows. 
First, univariate correlations $R(x_j,y)$ between each feature $x_j$ and the class label $y$ are computed, and only features with $|R(x_j,y)|$ above some threshold $\gamma$ are retained.
This yields a reduced feature matrix $\vc X_\gamma$, from which one then computes the first $n_c$ principal components $(z_1,\dots,z_{n_c})$ and uses these as the predictors for the reference model.
The advantage over the unsupervised principal components is that the screening step anticipates variation in the original features unrelated to the variance in $y$, and therefore the predictive power is typically more heavily loaded on the first few components.
In the experiments of this paper, the screening threshold $\gamma$ is selected using fivefold cross-validation from a coarse grid of $n_\gamma=7$ values evenly spaced between $\gamma_\tx{min}$ and $\gamma_\tx{max}$, where $\gamma_\tx{min}$ is the largest $\gamma$ such that none of the features are discarded and $\gamma_\tx{max}$ the smallest $\gamma$ such that only one feature survives the screening. 
Furthermore, we use $n_c=3$ SPCs with a Gaussian prior $\Normal{0,\tau^2}$ for the regression coefficients and hyperprior $\tau \sim \halfStudent[4]{0,s_\tx{max}^{-2}}$ where $s_\tx{max}$ denotes the standard deviation of the largest principal component (this is done only to make the prior roughly the same regardless of the scale of the SPCs).

We emphasize that we do \emph{not} argue that this gives a foolproof method for constructing a good reference model.
Rather the purpose is to demonstrate that even with such a simple, easy-to-implement and computationally light method it is possible to come up with a reference model that gives good results and improves feature selection in many cases.
Indeed, in our earlier work we found that the optimal method is in general data set dependent, and in some cases better results can be obtained by other choices such as the iterative version of the above algorithm~\citep{piironen2018}.
Again, cross-validation and posterior predictive checks should be used to guide the selection of the reference model~\citep{gelman2013book,vehtari2017b,gabry2018}.
A~generic strategy for improving the prediction accuracy is also to average over several models, using either stacking or (pseudo) Bayesian model averaging~\citep{yao2018}, or boosting or bagging in non-Bayesian context~\citep[see, e.g.,][]{hastie2009book}.

\section{Experiments}

This section presents several examples of the projective method. 
We shall first demonstrate the basic usage of the different projection techniques and the new LOO validation for the model size selection, and then compare the projective approach to the elastic net family estimators.
For fitting the Bayesian reference models we use Stan~\citep{stan_manual}, with the convenient interfaced to GLMs provided by R-packages {\tt rstanarm}~\citep{rstanarm} and {\tt brms}~\citep{burkner2017}.
All the projections are computed using our R-package {\tt projpred}.
The results for the elastic net family methods are computed using R-package {\tt glmnet}~\citep{friedman2010}.

\subsection{Illustration of different projections}
\label{sec:clusterdemo}

This section illustrates the differences between the three projection techniques introduced in Section~\ref{sec:projection_methods}.
Consider the following synthetic binary classification data. 
For instances belonging to the first class, the first three features are drawn from independent Gaussians with mean $1$ and scale $0.5$, whereas for the observations from the second class the mean and scale of these features are $-1$ and $0.5$, respectively.
In addition, the data has 27 additional noise features that are drawn from independent standard Gaussians, so the data has 30 features altogether (out of which the only the first three are predictive about the class label).
We generated one data realization with $n=50$ observations and fitted Bayesian logistic regression model to those data using the RHS prior~\eqref{eq:rhs_prior} with hyperparameter choices $p_0=1$, $s^2=1$ and $\nu=4$.
This serves as our reference model.
\begin{figure}[t]
	\centering
	\minput[pdf]{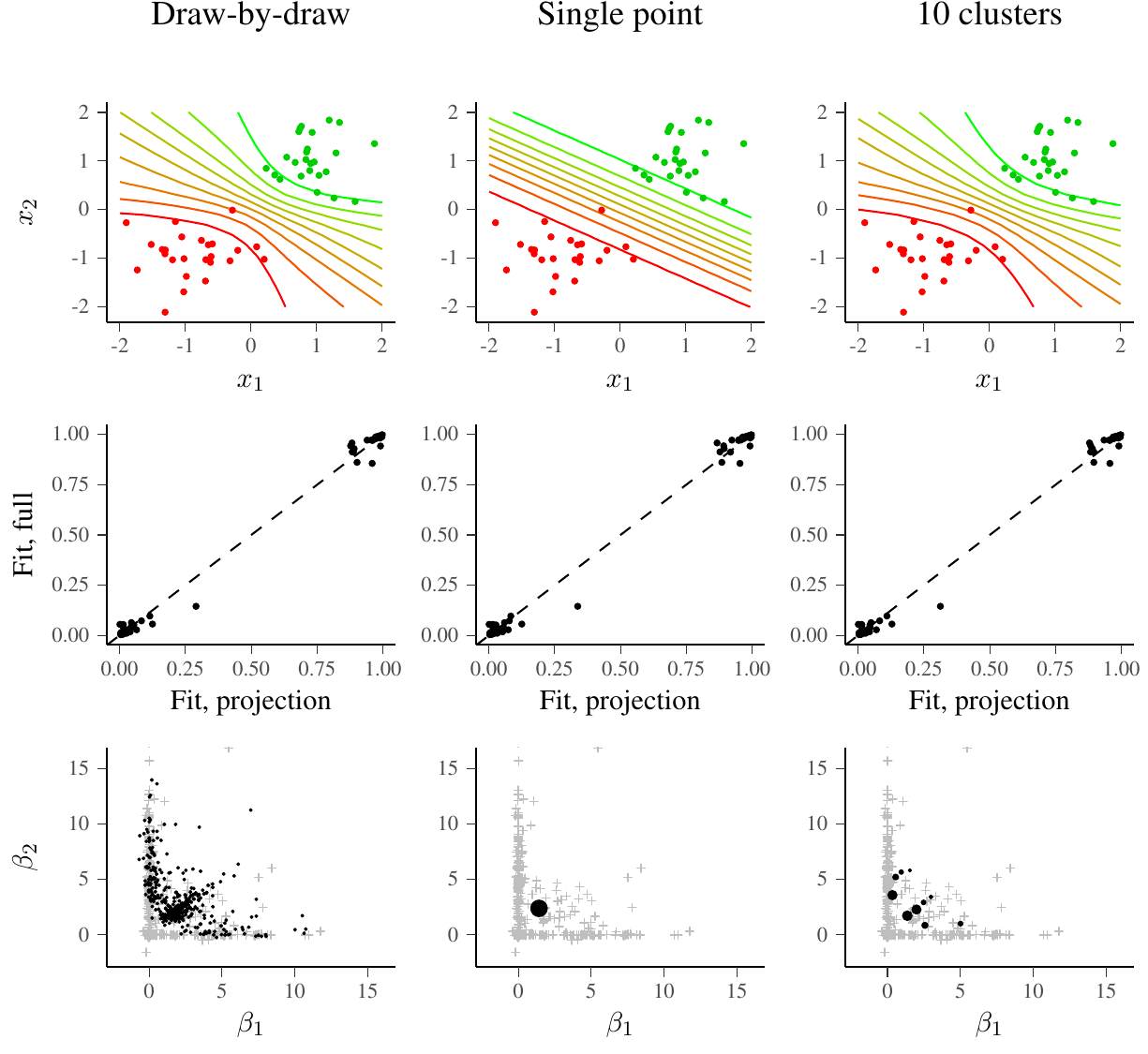}
	\caption{ {\it Demonstration of different projections:} The full posterior with $p=30$ features projected onto the first two features using the draw-by-draw approach (left column),  single point projection (middle column) or 10 clusters projection (right column). Top row shows the observed data and the contours (from 0.1 to 0.9) of the predictive probability for $\ti y = 1$, whereas middle row shows the predictive probabilities at the observed input locations (vertical axis denoting the result for the full model with all features, and horizontal axis for the projection of the corresponding column). Bottom row shows the projected regression coefficients (black dots) as well as the draws from the full posterior (gray crosses). In bottom row plots, the dot sizes denote the relative weights (the dot sizes between different columns are not comparable).} 
	\label{fig:clusterdemo}
\end{figure}

Figure~\ref{fig:clusterdemo} illustrates the posterior projection onto the first two features for the three different projections: draw-by-draw (left column), single point (middle column) and 10 clusters (right column).
We observe that even with the single point projection, the predictive probabilities are very close to those of the draw-by-draw projection (see top and middle row), and projecting 10 clusters gives predictions indistinguishable from the draw-by-draw projection for all practical purposes.
This result is insightful, as one might think that the single point projection would be substantially inferior because it computes only point estimates for the projected model.
The key insight is that these point estimates are computed so as to take into account the uncertainty in the parameters of the full model.
Therefore the resulting predictive distribution is much closer to that of the full model than what would be obtained by projecting only the posterior mean of the full model or by computing the maximum likelihood estimates for the submodel (which in this case do not even exist because the classes are separable).

Another important point is that even for the draw-by-draw method the projected posterior is in general different from the marginal posterior for those parameters in the full model (see the bottom left plot in Fig.~\ref{fig:clusterdemo}).
In particular, the projected posterior has vanishingly little mass near the origin $\beta_1=\beta_2=0$, although the full posterior has substantial mass there.
This makes sense: after removing feature $x_3$ which is predictive and highly correlated with $x_1$ and $x_2$ the coefficients of $x_1$ and $x_2$ can not \emph{both} be set to zero, otherwise the predictions would seriously be affected.

As discussed earlier, the benefit of the clustered projection compared to the draw-by-draw projection is its speed; projecting only $C$ clusters cuts down the computations by a factor of $C/S$, where $S$ is the number of draws that would be projected in the draw-by-draw projection.
The computational savings can be huge when projections need to be computed onto many models, such as with the LOO validation.
For instance, for this data set computing the projections of each of the $n=50$ LOO posteriors for all model sizes up to 30 features in a naive fashion would require a total of 1500 projections, each of which takes around a second or two depending on the hardware. 
Thereby with the clustered projection we can reduce the computation time from the order of 25--50 minutes to about 4--8 seconds\footnote{In a careful implementation the difference would not be quite as dramatic since some of the submodels would be visited in many of the $n=50$ folds, so their projections would not need to be computed again every time, but this example still gives a good idea of the computational gain.}.
The additional benefit of the single point projection is that it can be combined with the sparsity enforcing penalty functions (Sec.~\ref{sec:search_heuristics}) which allows for fast searching for promising submodels.

For these reasons, our preferred choice is to use one cluster projection in the selection phase, and a small number of clusters (such as 5--10) when making predictions with the submodels, especially if many submodels need to be considered.
Still, we find the draw-by-draw projection most convenient for visualizing the projected posterior distributions for instance when credible intervals or regions are of interest.
It also serves as a useful yardstick for checking and confirming the accuracy of the clustered projection.

\subsection{Simulated example revisited with projection and LOO}
\label{sec:toy_revisited}

We shall now revisit the simulated example discussed in Section~\ref{sec:toy_example} and illustrate the steps of projective selection as well as our new LOO validation technique.
The first step is to decide the reference model, which we would in practice do by assessing the fits of each of the candidate models using cross-validation and posterior predictive checks.
The sums of LOO log predictive densities for the Gaussian and RHS priors are $-76.8$ and  $-77.6$ with standard errors $6.8$ and $6.2$, respectively, so there is no significant difference between the predictive fits between these models (this holds also if we make the comparison in pairwise fashion, like in Eq.~\eqref{eq:rel_util_mean} and~\eqref{eq:rel_util_se}).
The R package {\tt spikeslab} does not provide the posterior draws for the regression coefficients and thus we cannot compute LOO for the SS prior, so we ignore it for now.

\begin{figure}[t]
	\centering
	\minput[pdf]{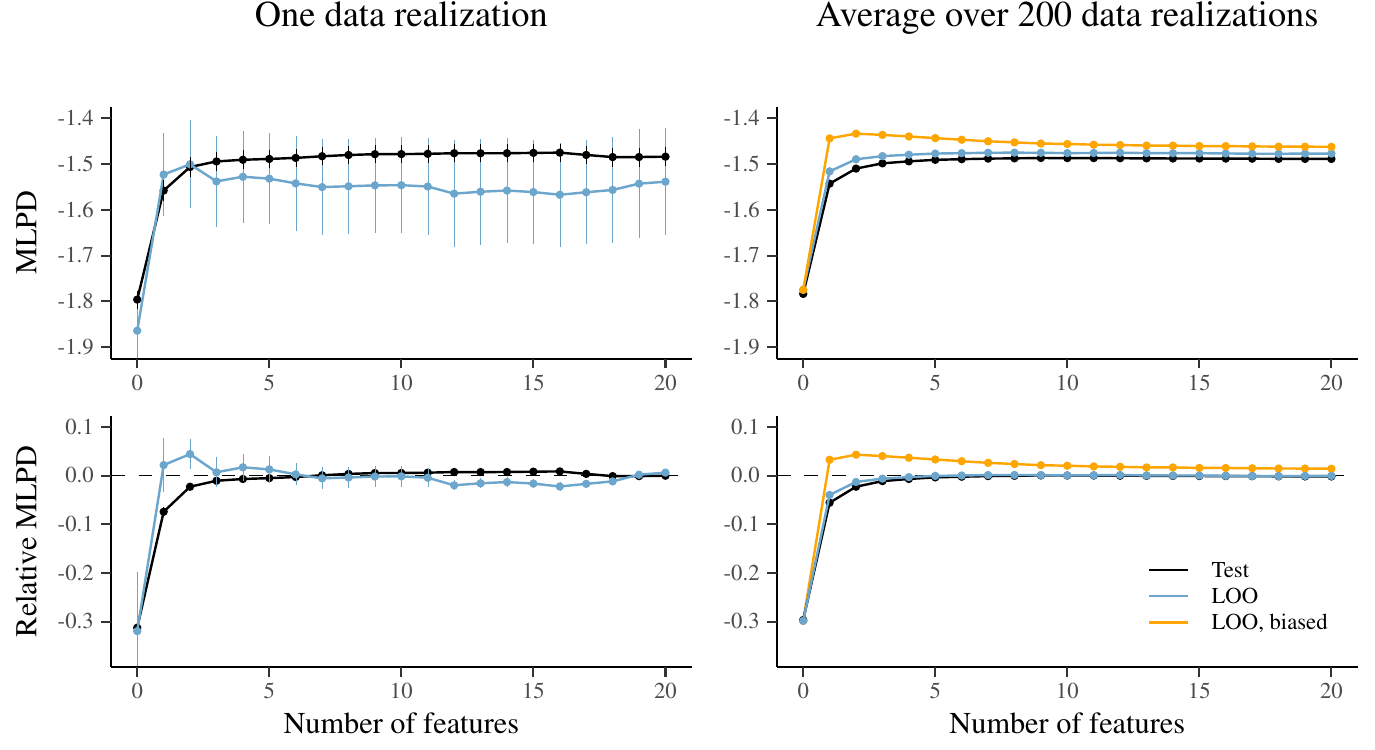}
	\caption{ {\it Simulated example, projective selection:} Left column: MLPD and relative MLPD with one standard error intervals on independent test data of 1000 points (black) and using LOO (blue) for the selected and projected submodels. The data has $p=50$ features and the reference fit is the linear model with RHS prior (the same as in the right middle subplot of Figure~\ref{fig:toy_marginals}). Right column: The same but results are averaged over 200 data realizations. The orange curves show the LOO for the submodels if the feature selection is done only once, and not separately for each of the $n$ folds. The difference to the blue curve comes from the selection induced bias. } 
	\label{fig:toy_assessment_rhs}
\end{figure}

Suppose we select the model with RHS prior as our reference model (the results for Gaussian prior are shown in Appendix~\ref{app:extra_results}).
We then run the projective feature selection with the $L_1$-search and assess the accuracy of the submodels using the LOO approach (Sec.~\ref{sec:loo_validation}).
The MLPD for the submodels relative to the reference model are shown in the bottom left subplot of Figure~\ref{fig:toy_assessment_rhs} (blue curve).
The one standard error -rule (Sec.~\ref{sec:kfold_validation}) would suggest selecting one feature, which results in a small loss in accuracy on test data (black curve) compared to the reference model.
The top left subplot shows the MLPD on the actual scale, which demonstrates how much larger the uncertainty is about the actual MLPD than about the relative MLPD.

The right column of Figure~\ref{fig:toy_assessment_rhs} shows the average LOO curves for both MLPD and relative MLPD over 200 data replications.
These graphs demonstrate that the actual LOO values for the submodels are slightly biased (optimistic).
This is due to a small bias in the PSIS-LOO for the reference model, which is also diagnosed by a few $\hat k$-values that exceed 0.7 in most data realizations.
Notice though, that the results for the relative MLPD are still essentially unbiased for submodels with performance close to the reference model, because the bias cancels out in subtraction~\eqref{eq:rel_util_mean} (see Sec.~\ref{sec:loo_validation}).
In other words, even if we have only a biased estimate of the reference model utility, we can still get a good indication of whether our submodel performance is close to that of the reference model.

Right column subplots of Figure~\ref{fig:toy_assessment_rhs} also show the expected LOO results if we do \emph{not} take into account the selection induced bias but perform the selection only once (not separately for the $n$ folds) and then compute LOO for the submodels (see Sec.~\ref{sec:search_validation}).
The selection induced bias is clear although only moderate in this particular example.

For assessing the submodel accuracies, LOO validation is very useful in this particular example because of a few reasons.
Firstly, PSIS-LOO works pretty well for the full model (only a few $k$-values above 0.7 in most data realizations).
Secondly, the number of features is only moderate and hence the feature selection is very fast. 
Thirdly, the number of observations is small, so the number of selection paths we need to compute is also small.
Consequently, the whole validation process takes only a few seconds, which is much less than a single model fit with the horseshoe prior (around half a minute with a standard laptop), so the computational savings compared to $K$-fold cross-validation are clear.

\subsection{The benefit of using a reference model}
\label{sec:benefit_of_reference}

This section demonstrates the benefits of a reference model for feature selection and parameter estimation in the submodels.
We again utilize simulated data generated by mechanism~\eqref{eq:toy_data}, and consider both regression with the original $y$ regressed on $(x_1,\dots,x_p)$, and binary classification with target variable defined as an indicator $y_\tx{class} = \indicator{y>0}$.

We used a setup with $n=50$ training observations with $p=500$ features, out of which first $p_\tx{rel}=50$ were relevant, and report average results over 50 data realizations for $\rho$-values of $0.3$, $0.5$ and $0.8$.\footnote{We also considered varying values for $p_\tx{rel}$ but the conclusions are not sensitive to the selected value $p_\tx{rel}=50$.}
The reference model is fitted using SPCs as discussed in Section~\ref{sec:refmodel_construction}. 
We tested four different strategies for selecting features and making predictions with the selected subsets of features:
\begin{enumerate}
	\item \emph{Lasso}: Sort the variables from the most relevant to least relevant according to the order in which they enter the model as the regularization coefficient $\lambda$ is decreased. For a given number of features, the submodel coefficients are computed using the smallest~$\lambda$ for which other variables do not enter the model. In the regression problems, the noise variance~$\sigma^2$ is estimated as proposed by \cite{reid2016}, that is, by dividing the sum of the squared residuals by $n-p_\tx{act}$ where $p_\tx{act}$ denotes the number of active features in the submodel.
	\item \emph{Lasso, relaxed}: Same as Lasso, but after sorting the variables, the submodel coefficients and predictions are computed without any regularization (which affects also the estimated noise variance in regression).\footnote{We are aware that the term `relaxed Lasso' has been used to denote a more general method where after feature selection the coefficients are computed with a small but nonzero $L_1$-penalty~\citep{meinshausen2007}. The complete relaxation (i.e., zero penalty after selection) was referred to as `Lasso-OLS hybrid' by \cite{efron2004}}
	\item \emph{$L_1$-projection}: $L_1$-penalized projection~\eqref{eq:elnet_projection} varying $\lambda$  similarly as in Lasso. In regression, the projected noise variance is computed according to Equation~\eqref{eq:proj_sigma2} (where $C=1$).
	\item \emph{$L_1$-projection, relaxed}: Same as $L_1$-projection, but after sorting the variables, the submodel coefficients are projected without any regularization (which affects also the projected noise variance in regression).
\end{enumerate}
Notice that all these methods utilize only point estimates for the model parameters in the submodels, the difference is only how they are computed.

\begin{figure}[t]
	\centering
	\minput[pdf]{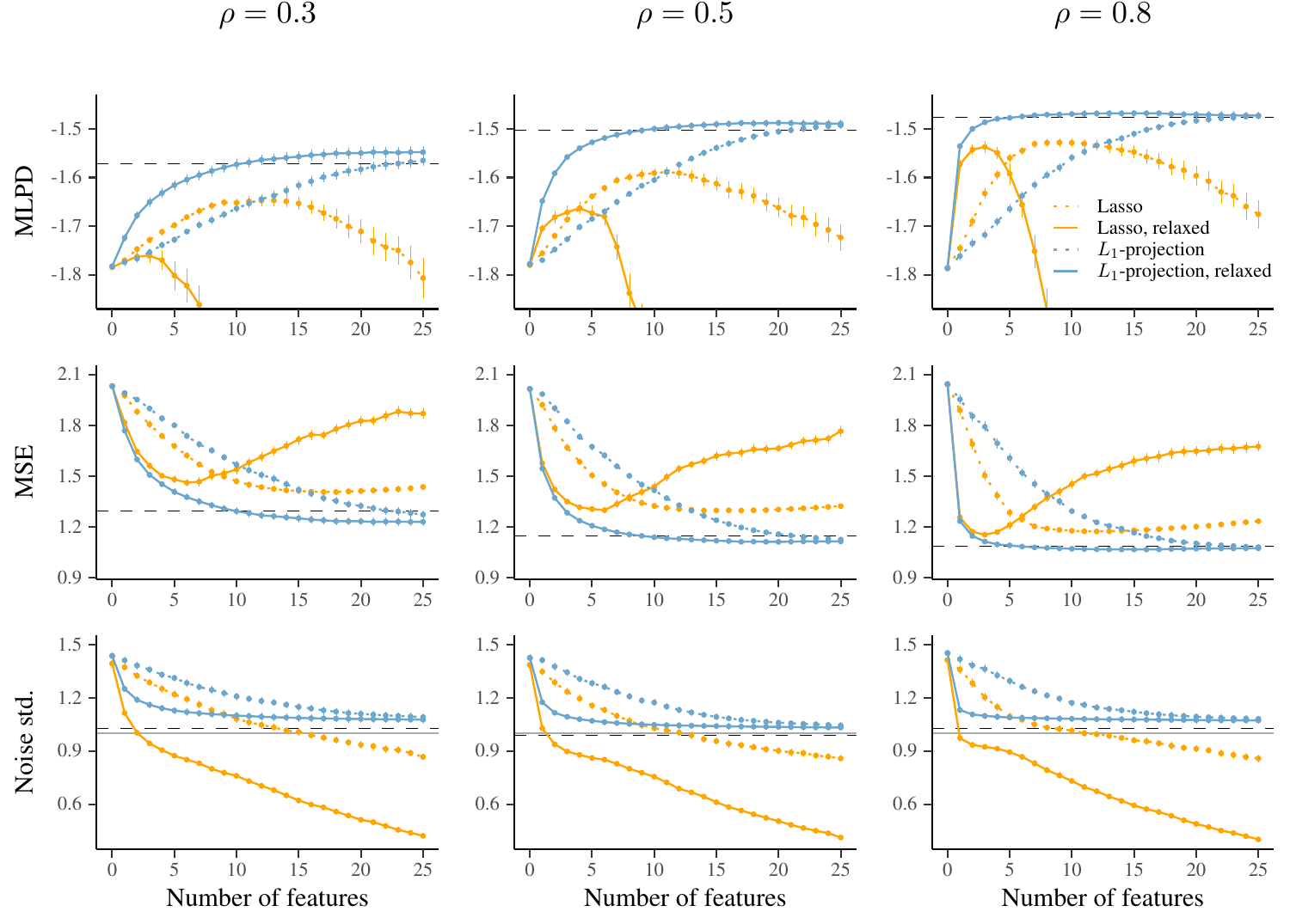}
	\captionspace
	\caption{ {\it Benefit of reference model, regression:}  MLPD and MSE on test data, along with the estimated noise standard deviation as a function of number of features selected after $L_1$-penalized search, before and after relaxation (dashed and solid, respectively), with and without utilizing the reference model (blue and orange, respectively). Different columns show results for different values of $\rho$ (see Eq.~\eqref{eq:toy_data}). Errorbars indicate one standard error intervals and black dashed lines the reference model result. In the bottom row plots the gray line denotes the true noise standard deviation.} 
	\label{fig:proj_vs_lasso_reg}
\end{figure}
\begin{figure}[t]
	\centering
	\minput[pdf]{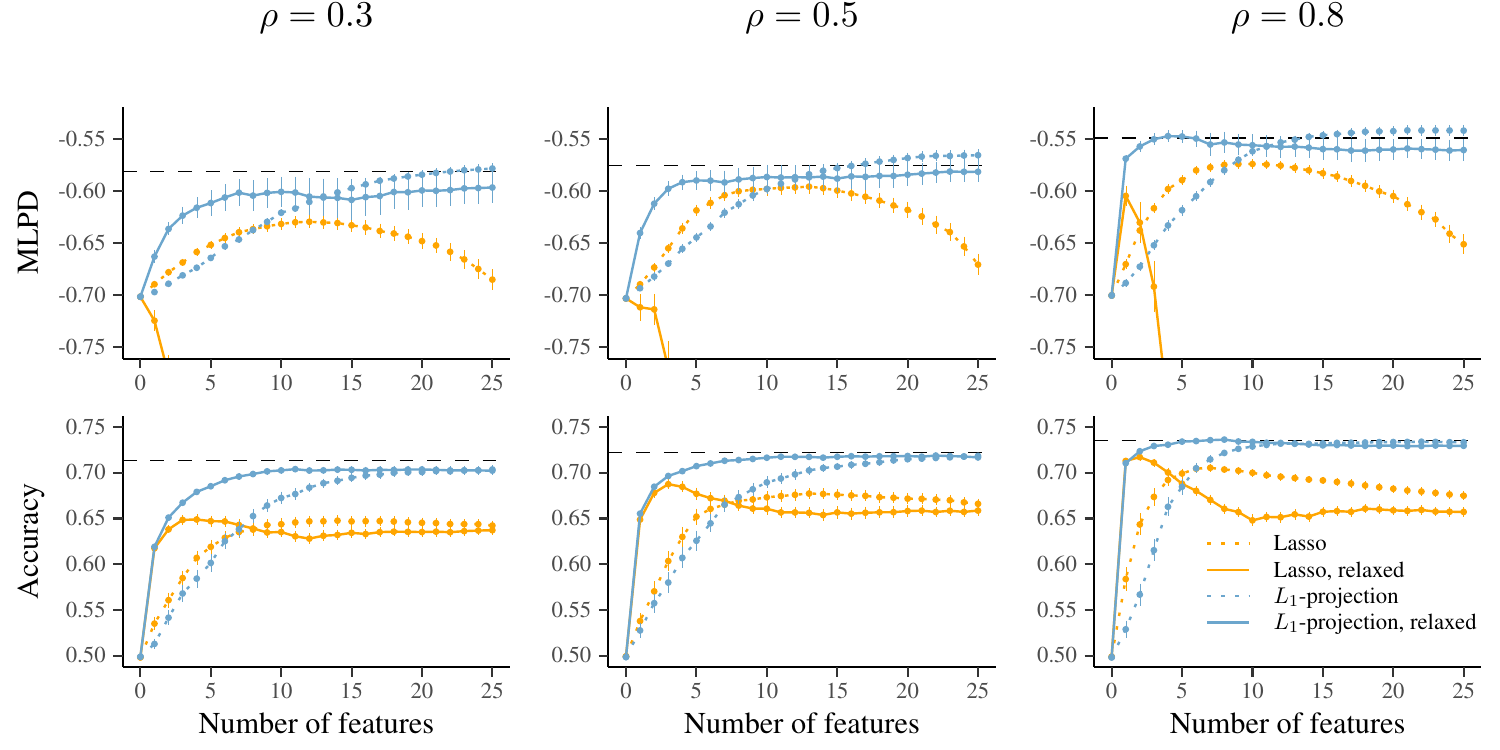}
	\captionspace
	\caption{ {\it Benefit of reference model, classification:} Results analogous to those in Figure~\ref{fig:proj_vs_lasso_reg} but for the classification data. Here shown are MLPD and classification accuracy on test data.} 
	\label{fig:proj_vs_lasso_classif}
\end{figure}

Figure~\ref{fig:proj_vs_lasso_reg} shows the regression MLPD and MSE on independent test data as well as the projected noise standard deviation for different submodel sizes.
The blue curves demonstrate the benefit of relaxation for $L_1$-projection: both eventually achieve the performance of the reference model but without relaxation this requires many more features.
The reason is the inherent tradeoff between shrinkage and selection: in order to force most of the regression coefficients to zero, the regularization coefficient $\lambda$ must be made large, but this will also overshrink the nonzero coefficients.
Therefore projecting without any penalization after selection achieves greatly improved tradeoff between accuracy and model complexity.
Notice in particular that here no regularization is needed to avoid overfitting in projection; when more features are added the projected submodels simply get closer to the reference model.

However, the picture is quite different when the parameter estimates are computed based on the observed data without utilizing the reference model (Fig.~\ref{fig:proj_vs_lasso_reg}, orange curves).
The relaxation improves the fit in terms of MSE for submodels with only a few features but results in overfitting for larger models.
In terms of MLPD the relaxed Lasso performs worst overall indicating badly calibrated uncertainties in the predictive distributions, which is mostly due to underestimation of the noise variance (bottom row) for most model sizes.
Projection methods on the other hand show very good calibration of predictive uncertainties which is evident from superior MLPD and noise variance estimation for most model sizes.
Overall the projection approach shows a bigger edge for~$\rho=0.3$ and $\rho=0.5$ where the individual features are less predictive.

Figure~\ref{fig:proj_vs_lasso_classif} shows the analogous results for the classification data.
The conclusions are very similar to those drawn from the regression example.
Here the relaxed Lasso overfits even more severely; although the classification accuracy is similar to the Lasso, the MLPD is very low indicating bad calibration in the predicted class probabilities.
Again, the edge for projection is more pronounced for $\rho=0.3$ and $\rho=0.5$, but we observe that for these cases also the relaxed projection struggles to achieve the same MLPD as the reference model, and for larger number of features (15--25) the penalized projection achieves slightly better results.
This is due to a small instability of the projection in data sets where some of the reference class probabilities are close to 0 and 1 and shows that even the projection, although very resilient, is not always entirely immune to overfitting.

\subsection{Real world benchmarks}
\label{sec:microarray_data}

This section shows how the projection compares in high-dimensional real world problems.
We use microarray data sets\footnote{All except the Ovarian data are available at \url{http://featureselection.asu.edu/datasets.php}.} some of which have been used as benchmarks by several authors~\citep{li2002,lee2003,hernandezlobato2010}.
All data sets deal with binary classification, and the number of features and data set sizes can be found in Table~\ref{tab:computation_times}. 

Again, as a reference model we use the one described in Section~\ref{sec:refmodel_construction} and call it here `Bayes SPC'.
For the projection method, we used $L_1$-search and made the submodel predictions using five clusters projection.
Notice that although the reference model employed only a reduced set of features (those that survived the screening), the projective selection considered all the features (no pre-selection).
The number of features was decided based on fivefold cross-validation.
To investigate the effect of the model size selection heuristic discussed in Section~\ref{sec:kfold_validation}, we report results for the smallest number of features that had its cross-validated MLPD within one standard error away from the reference model (`Proj-ref-1se') or from the best submodel ('Proj-best-1se').
We also report results ('Proj-ref-1se-reg' and 'Proj-best-1se-reg') that are otherwise exactly the same as the two above but utilize a little bit of ridge regularization (with $\lambda=0.1$) in the submodel projections which was observed to improve the numerical stability in cases where some of the reference model class probabilities are close to~0 and~1.

For comparison, we computed results for Lasso, elastic net (with $\alpha=0.7$ and $\alpha=0.3$) and ridge.
To investigate the sensitivity of these to the selection of the regularization parameter $\lambda$, we report results for two choices: $\lambda_\tx{opt}$ denotes the value that minimizes the tenfold CV-error whereas $\lambda_\tx{1se}$ (default in {\tt glmnet}) denotes the largest $\lambda$ which has its CV-error within one standard error of that of~$\lambda_\tx{opt}$.
To avoid any possible biases in the comparisons, the out-of-sample predictive accuracies for all the methods were assessed using an outer tenfold cross-validation.
That is, the reference model, projected submodels as well as the baseline methods were computed ten times, each time leaving one tenth of the data out and then validating the found models on this left-out data.

\begin{figure}%
\vspace{-0.99cm}
	\centering
	\minput[pdf]{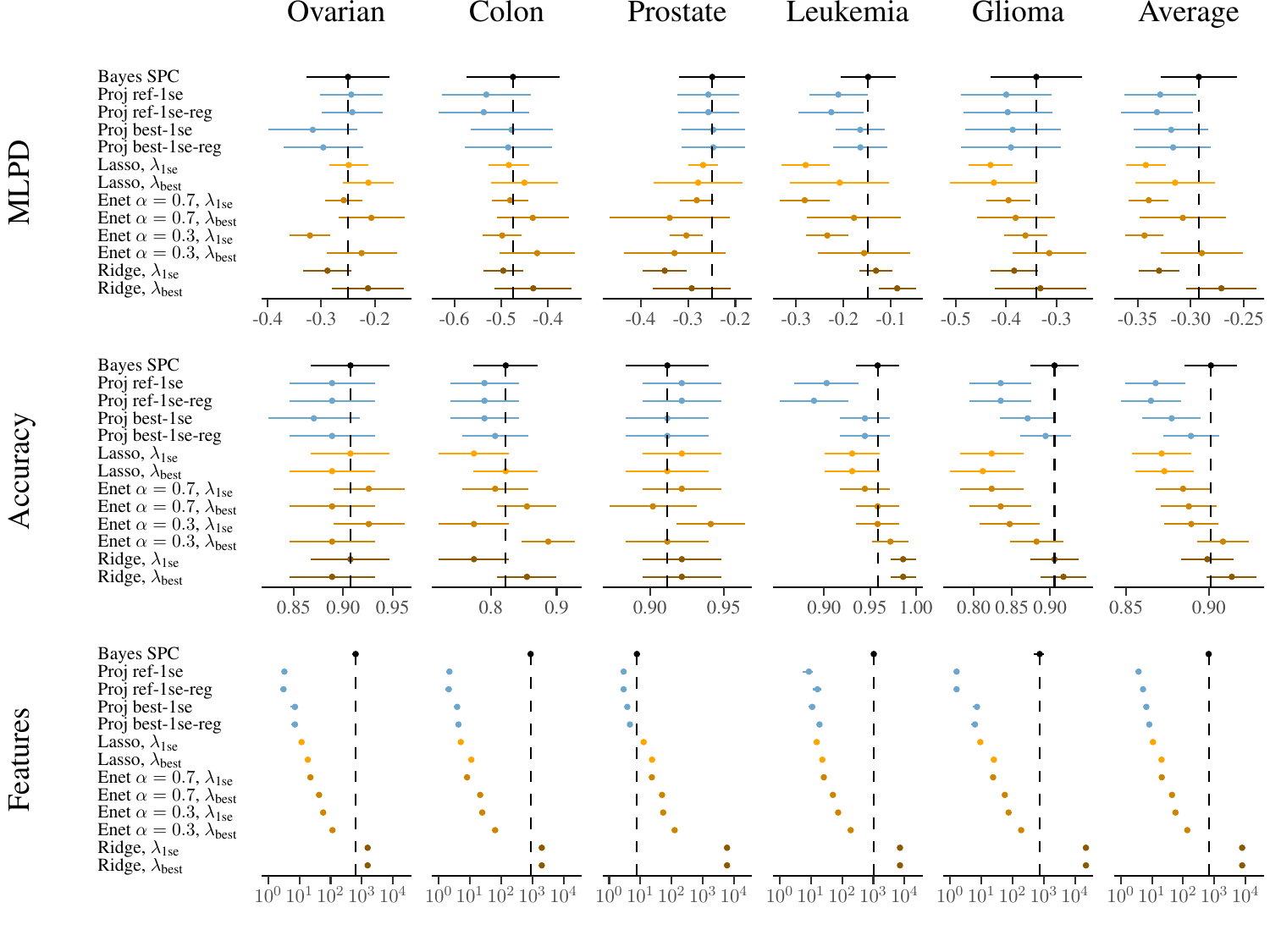}
	\vspace{-1cm}
	\caption{ {\it Microarray benchmark data sets:} MLPD (top row), classification accuracy (middle row) and the number of features used (bottom row) with one standard error intervals for the different data sets. The last column denotes the average. In all plots the dashed vertical line denotes the results for the Bayes SPC that is used as the reference for the projections. Many methods produce comparable predictive accuracy but the projection methods achieve the same accuracy with far fewer features (notice log scale in the bottom row plots). }
	\label{fig:realdata}
\end{figure}

The MLPD and classification accuracies from the outer cross-validation are shown in the first two rows of Figure~\ref{fig:realdata}.
Overall the differences between the methods are fairly small compared to the standard errors in the estimates.
In terms of MLPD, the reference model Bayes SPC gives somewhat better results than Lasso, elastic net and ridge with $\lambda_\tx{1se}$, but all these give similar results when $\lambda_\tx{opt}$ is used, and in fact ridge gives a bit better results for Leukemia data.
All projections have statistically indistinguishable MLPD compared to Bayes SPC, but the model size selection with `best-1se' performs slightly better in terms of classification accuracy.
Adding a little bit of regularization does not hurt predictive accuracy but we noticed that it makes the projection numerically more stable in cases where the reference class probabilities are close to 0 and 1.

The bottom row of Figure~\ref{fig:realdata} shows the number of selected features for each method.
The projection methods produce by far the most parsimonious models (notice the log scale).
The only data set where Lasso (with $\lambda_\tx{1se}$) selects fewer variables is Leukemia, but there it also yields inferior results in terms of MLPD.
This is perfectly in accordance with the results shown in Figures~\ref{fig:proj_vs_lasso_reg} and~\ref{fig:proj_vs_lasso_classif}: the projection finds very good tradeoff between sparsity and accuracy.
To fully respect the differences in the number of features used, we have also reported them using hard numbers in Table~\ref{tab:nfeat} (appendix~\ref{app:extra_results}) since an accurate comparison on the log scale is somewhat cumbersome.

The computation times are shown in Table~\ref{tab:computation_times}.
After forming the reference model, the projection is computationally only slightly more expensive than Lasso and the increase comes from the relaxed projections (the predictions are computed without the $L_1$-penalty).
Although not as highly optimized as {\tt glmnet}, our software is reasonably fast even for the largest problems.
Forming the reference model (Bayes SPC) is computationally the most expensive part, though still very affordable considering that all the computations (reference model construction and projection) for the largest number of features can be done in about two minutes.
Indeed, this demonstrates that the projection can be very feasible computationally and it can yield improved results to the standard approaches, as were shown in Figure~\ref{fig:realdata}.

\section{Theoretical results}

In this section we present a theorem that helps us to understand when the reference model could be helpful for parameter learning in linear submodels.
Here we only state the results, the proofs can be found in appendix~\ref{app:proofs}.

\begin{table}%
\vspace{-0.1cm}
\centering
\abovetopsep=2pt
\begin{tabular}{ lccrrrrr }
\toprule
Data set & $n$ & $p$ & \multicolumn{5}{c}{Computation time} \\
\cmidrule(r){4-8}
 & & & Bayes SPC & Projection & Lasso  ({\tt \small projpred}) & Lasso  & Ridge \\ 
\midrule
Ovarian & 54 & 1536 & 30.4 & 3.6 & 1.3 & 0.2 & 1.5 \\
Colon & 62 & 2000 & 31.0 & 4.0 & 1.6 & 0.3 & 2.2 \\
Prostate & 102 & 5966 & 49.4 & 7.6 & 5.0 & 0.8 & 7.5 \\
Leukemia & 72 & 7129 & 47.0 & 6.3 & 5.6 & 0.7 & 9.4 \\
Glioma & 85 & 22283 & 95.8 & 14.2 & 15.6 & 2.6 & 52.2 \\
\bottomrule
\end{tabular}
\caption{{\it Microarray benchmark data sets:} Average computation time (in seconds) over five repeated runs. In all cases the time contains the cross-validation of the tuning parameters and/or the model size. The first result for Lasso is computed using our software ({\tt projpred}) whereas the second result (and that of ridge) is computed using the R-package {\tt glmnet} which is more highly optimized. }
\label{tab:computation_times}
\end{table}

Let $\vc X = (\vc x_1^\tp,\dots, \vc x_n^\tp) \in \Re^{n \times p}$ be the design matrix and $\vc y = (y_1, \dots, y_n) \in \Re^n$ the target measurements. 
Assume the target measurements decompose as $y_i = \mu(\vc x_i) + \varepsilon_i$, where $\mu(\vc x)$ is the true expected value of $y$\, given $\vc x$, $\mu(\vc x) = \mean{y \given \vc x}$, and $\varepsilon_i$ are i.i.d. random numbers independent of~$\vc x$ with zero mean and finite variance $\sigma^2$ denoting the variation in $y$ that cannot be explained by~$\vc x$.
It should be emphasized that although we assume $\varepsilon$ denotes random error independent of $\vc x$, it may contain systematic variation related to some other (unobserved) features not included in $\vc x$, and hence the magnitude of $\varepsilon$ should be interpreted as the irremovable error for this particular set of features $\vc x$.
In vector notation, $\vc y$ decomposes as $\vc y = \vs \mu + \vs \varepsilon$, where $\vs \mu = (\mu(\vc x_1),\dots,\mu(\vc x_n))$ and $\vs \varepsilon = (\varepsilon_1,\dots,\varepsilon_n)$.
Furthermore, in what follows we shall use the shorthand notation $||\vc v||^2_{\vc M} = \vc v^\tp \vc M \vc v$, where $\vc v$ is a vector and $\vc M$ a positive definite matrix.

Consider two methods of estimating the regression coefficients when regressing $\vc y$ on $\vc X$, namely 
\begin{align}
	\vs {\hat \beta} = (\vc X^\tp \vc X)^{-1} \vc X^\tp \vc y \quad \tx{and} \quad
	\vs \beta_\perp = (\vc X^\tp \vc X)^{-1} \vc X^\tp \vs \mu_*.
\label{eq:coeff_ls_and_ref}
\end{align}
Here $\vs{\hat \beta}$ is the familiar least squares estimate, and $\vs \beta_\perp$ a projection of an arbitrary reference fit $\vs \mu_* \in \Re^n$.
Let us then define the expected prediction error for any vector of coefficients $\vs \beta$  as
\begin{align*}
	\Delta(\vs \beta) = \mean[\vc{\ti y}]{ \frac{1}{n}|| \vc X \vs \beta  - \vc{\ti y} ||^2 }.
\end{align*}
Notice that although we consider here the predictions at the observed input locations $\vc X$, the expectation is with respect to a set of \emph{new} measurements $\vc{\ti y} = \vs \mu + \vc {\ti e}$, where $\vc{\ti e}$ is a vector of new noise terms $\ti \varepsilon_1,\dots,\ti \varepsilon_n$.
The gain from using $\vs \beta_\perp$ instead of $\vs{\hat \beta}$ is defined as the reduction in the expected prediction error
\begin{align}
	G = \Delta(\vs{\hat \beta}) - \Delta(\vs \beta_\perp). 
\label{eq:gain}
\end{align}
We have the following lemma.
\begin{lemma}
Assume regression coefficient estimators $\vs{\hat \beta}$ and $\vs \beta_\perp$ as defined by Equation~\eqref{eq:coeff_ls_and_ref}. The gain~$G$ (Eq.~\eqref{eq:gain}) of using $\vs \beta_\perp$ instead of $\vs{\hat \beta}$ satisfies 
\begin{align*}
	G = \frac{1}{n}\left( ||\vc y - \vs \mu||_{\vc P}^2 - ||\vs \mu_* - \vs \mu||_{\vc P}^2 \right),
\end{align*}
where $\vc P = \vc X (\vc X^\tp \vc X)^{-1} \vc X^\tp$.
\label{lemma:gain}
\end{lemma}
Since both $||\vc y - \vs \mu||_{\vc P}^2$ and $||\vs \mu_* - \vs \mu||_{ \vc P}^2$ are non-negative, the interpretation of Lemma~\ref{lemma:gain} is that for linear submodels, one can expect to gain (that is, $G \ge 0$) from using a reference model when the reference fit $\vs \mu_*$ is closer to the best possible prediction $\vs \mu$ (with features $\vc x$) than the observed noisy target values $\vc y$ (with the norms taken with respect to the projection matrix $\vc P$).
This makes perfect sense: if we fit our model to pseudo-data $\vs \mu_*$ instead of the actual data $\vc y$, we expect to do better if the pseudo-data are closer to the true underlying conditional mean $\mu(\vc x) = \mean{y\given \vc x}$, that is, less noisy than the actual data.
Notice that the lemma makes no assumptions about the form of the true underlying mean $\mu(\vc x)$ that captures the relationship between $y$ and~$\vc x$.
In particular, $\mu(\vc x)$ need not be linear in $\vc x$, not even smooth or continuous.
Neither does the lemma assume anything about how the reference fit $\vs \mu_*$ is constructed.

Let us now assume the differences $\vc e_* = \vs \mu_* - \vs \mu$ are random numbers with mean $\vc b$ and covariance~$\vc K$.
These describe the bias and variance in the reference fit.\footnote{Here we mean bias and variance both due to fitting the reference model to a finite data set and having unobserved features. For example, even if our reference model was a completely deterministic function of $\vc x$ and some other features $\vc z$, then its value in a particular location $\vc x_i$ is still random as it depends on the realized value for $\vc z$.}
We have the following theorem
\begin{theorem}
Assume the terms $\vc e_* = \vs \mu_* - \vs \mu$ have mean $\vc b \in \Re^n$ and covariance $\vc K \in \Re^{n\times n}$. Then the expected gain can be written as
\begin{align*}
	\mean{G} = \frac{1}{n}\left( \sigma^2 p - \tr{\vc P \vc K} - ||\vc b||_{\vc P}^2 \right).
\end{align*}
\label{thm:gain}
\end{theorem}
Theorem~\ref{thm:gain} further decomposes the reference model error into bias and variance.
The term $\tr{\vc P \vc K}$ is difficult to grasp without further assumptions, but the theorem can be understood more easily by the following immediate corollary.
\begin{corollary}
Assume the reference model errors are uncorrelated with a common variance, that is, $\vc K = \sigma_{\mu_*}^2 \vc I$.
Then the expected gain $\mean{G}$ simplifies to
\begin{align*}
	\mean{G} = \frac{p}{n} \left( \sigma^2  - \sigma_{\mu_*}^2   - \frac{1}{p} ||\vc b||_{\vc P}^2 \right).
\end{align*}
\label{clry:uncor_err}
\end{corollary}
This corollary states that with an unbiased reference model ($\vc b = 0$)  we can expect to gain when the variance of the residuals $\vs \mu_*-\vs \mu$ is smaller than the variance of $\vc y - \vs \mu$.
Furthermore, the gain increases with the dimensionality of the projection space $p$, but on the other hand goes to zero when $n \rightarrow \infty$.
This is also in perfect accordance with the empirical results, for instance those shown in the middle row of Figure~\ref{fig:proj_vs_lasso_reg}.
There the difference in predictive MSE between relaxed Lasso and projection is small up to about $p=2$, but then starts to increase gradually.
On the other hand we know the least squares fit gives us the optimal coefficients at the limit $n \rightarrow \infty$ and hence we do not expect to gain anything then with a reference model.

The above analysis assumes the future predictions are made at the observed input locations.
Usually this is not quite a realistic assumption, but it still gives us some idea when the reference model could be useful.
Extending the result to different $\vc {\ti x}$ would require assumptions about the functional form of $\mu(\vc x)$.
Furthermore, here we used squared error as the loss function as it allows for tractable analysis, but we expect one of the major advantages of the projection to be that it conserves the predictive uncertainties well which are not measured by the squared error.
Finally, this analysis considers only parameter learning in the submodels but it says little about when the reference model can improve the \emph{selection} of a better feature combination.
In principle it is possible to improve selection even when the reference model is not unbiased, as long as the bias is ``in the right direction'', for instance so that it favors certain features over the others.
We have discussed in a bit more detail in our earlier work, see Section~3 in \cite{piironen2017a}.
The empirical evidence about the improved selection is convincing (Sec.~\ref{sec:intro_example}) but currently we are not aware of any theoretical analysis on this topic.

\section{Discussion}

Below we provide some final remarks, together with recommendations for practitioners and possible directions for future research.

\subsection{Projection versus Lasso and related methods}

Although our results indicate that with a reasonable reference model the relaxed $L_1$-projection outperforms Lasso in terms of tradeoff between sparsity and accuracy, there is no question that Lasso and the whole elastic net family are useful methods.
These methods are fast to fit and often provide good results without any hand tuning, so they are very useful for getting a good predictive baseline for almost any problem.

In addition to better tradeoff between sparsity and accuracy, the projection has the upside that it provides a principled approach to handling also parameters other than the regression coefficients.
For instance, projecting the noise variance in linear regression is trivial and analytical solution exists (see Sec.~\ref{sec:glm}) whereas for Lasso this is considered to be a difficult problem~\citep{reid2016}.
This property is very beneficial in all settings where additional dispersion parameters need to be estimated.
These situations come up frequently, such as in survival analysis with parametric observation models~\citep{peltola2014}.
Another benefit is that the projection gives a principled answer to how to make predictions when some of the features we would like to use are unavailable at prediction time (missing data)~\citep{lindley1968}.

\subsection{Multiple hypothesis testing}
\label{sec:multiple_hypothesis}

In this paper we have focused on selecting a minimal subset of features that are sufficient for achieving accurate predictions.
As pointed out in Section~\ref{sec:selection_terminology}, this is a different problem from what is known as multiple hypothesis testing, where one is less interested in predictive accuracy but the desire is to identify all the features that are statistically related to the target variable.
For instance, in gene expression studies it is conventional to try to identify genes whose expressions are different between two groups (say between control and cancer patients).
This is often done by computing some statistic for each feature (say two sample $t$-statistic) and then based on these trying to distinguish between significant and non-significant features using either frequentist of Bayesian approaches.
For a nice overview, see \cite{efron2010book}, and for more recent Bayesian accounts, see \cite{bhattacharya2015,bhadra2017} and \cite{vanderpas2017}.
The empirical evidence indicates that the reference model approach could be highly useful also in this problem setting since it tends to help rank the truly relevant features before the irrelevant ones (Sec.~\ref{sec:intro_example}).
Still, it is an open question which approach to use for detecting the truly relevant features, but this is a research area on its own.

\subsection{Future directions}

There are several natural directions to continue this work.
An obvious topic for generalized linear models would be implementing the projection to multiclass classification which has not been done yet but would be straightforward.
Another useful noise model would Student-$t$ which can find applications in data sets with outliers.
This likelihood is not log-concave, but the minimization of the KL-divergence could be implemented using the EM-approach that can be used to find the maximum likelihood solution for the regression coefficients.
Other relatively straightforward extensions would be hierarchical GLMs and generalized additive models (GAM) which provide more flexibility but for which the projection should be implementable using the methodology presented in Section~\ref{sec:projection}.

Ultimately we would like to extend the projection framework to nonlinear models such as Gaussian processes (GP).
This topic was tentatively pursued in \cite{piironen2016} with promising results, but that approach is computationally too expensive and prohibitive for large data sets.
It could be possible to formulate the projection in a more clever way by borrowing ideas from the stochastic variational inference algorithms that have proven useful for GP learning with large data sets.
Another difficulty with GPs is that due to their flexibility, the minimization of KL-divergence locally at training inputs does not necessarily guarantee small divergence elsewhere.

We also believe the projective approach could find more applications in improving interpretability and transparency of complex black box models such as deep neural networks, an idea that was explored recently by \cite{ribeiro2016} and \cite{peltola2018}.
For example, in image classification one could approximate the prediction surface of the complex classifier with a linear model in the vicinity of a misclassified image to figure out which of the pixels had high weight in making the decision.

\acks{ 
We thank Michael Riis Andersen for helpful discussions.
}

\begin{appendices}

\section{Subsampling LOO}
\label{app:subsampling_loo}

As discussed in Section~\ref{sec:subsampling_loo}, one approach to perform validation for large $n$ is to use only a subset of $m<n$ points in LOO validation.
This gives an unbiased utility estimate for each submodel but with higher variance than if we used all $n$ LOOs.

The variance can be reduced by a semi-random subsampling.
The PSIS procedure does not only smooth the importance weights but gives also an indication of the tail thickness of the importance weight distribution, measured by the $\hat k$-value for each of the $n$ datapoints.
The larger the $\hat k$-value, the fatter the tail indicating more influential data point.
In particular, if $k \le 0.5$, the raw importance weights have a finite variance, central limit theorem kicks in and the Monte Carlo error decreases quickly. For 
$0.5 < k \le 1$ the raw weights have infinite variance, but generalized central limit theorem holds and the importance sampling estimate is asymptotically consistent. However, pre-asymptotic behavior is such that for $k>0.7$ infeasible high sample sizes would be needed for reasonable error rates. For $k<0.7$ the Pareto smoothing regularizes the estimate so that it has a finite variance with a cost of small bias, and empirical results indicate practically useful convergence rate \citep[see][for more thorough discussion]{vehtari2017b,vehtari2017a}.
We divide the data points into 3 strata: those with $\hat k < 0.5$ (good), those with $0.5 < \hat k < 0.7$ (OK) and those with $\hat k > 0.7$ (bad).
Denote the sizes of these strata as $n_1$, $n_2$ and $n_3$, respectively.
We then randomly draw $\min(m/3, n_j)$ points without replacement from each stratum $j=1,2,3$.
If the number of drawn points is less than $m$, the rest of the points are drawn randomly from the remaining points so that $m$ points are selected in total.
Let $m_j, \,j=1,2,3$ denote the number of points drawn from each stratum.
To account for the fact that sizes of the strata are different, the points are weighted according to the stratum sizes when computing the utility estimates.
Jumping straight to the result, the point estimate and its standard error for the quantity $\Delta U_k$ (analogous to Equations~\eqref{eq:rel_util_mean} and~\eqref{eq:rel_util_se}) are given by
\begin{align}
	\Delta \bar U_k &= \sum_{i=1}^n v_i \left( u_k^{(i)} - u_*^{(i)} \right), 
	\label{eq:rel_util_mean_ss} \\
	s_k &= \sqrt{\frac{1}{m} \svar[i=1]{n}{u_k^{(i)} - u_*^{(i)}, v_i}},
	\label{eq:rel_util_se_ss}
\end{align}
where $\svar[i=1]{n}{\, \cdot, v_i}$ denotes the weighted sample variance with weights $v_i$.
Each of the $m_1$ points in the `good' category is given weight $ n_1/(n m_1)$, and the weights for the points in the `ok' and `bad' categories are $n_2/(n m_2)$ and $n_3/(n m_3)$, respectively.
The points that did not get selected in the subsample are assigned zero weight $v_i = 0$.

The standard error~\eqref{eq:rel_util_se_ss} will be higher than~\eqref{eq:rel_util_se}, but often a fairly small number of points, such as $m=200$ is enough for obtaining reasonably accurate estimates of the difference between the reference and submodels.
This is because the submodels become closer to the reference model as more features are added, and therefore also the differences between their pointwise utilities become ever smaller and hence the standard errors for the models with good performance can get small already for small number of LOOs.
It is worth noting though, that although the standard error for the utility difference would be small, the standard error for the {\it actual} utility $U_k$ can be substantial for small $m$.
If all LOOs are computed for the reference model, it is possible to combine the standard errors for $U_*$ and $U_k'$ to get a smaller standard error for $U_k$ but we shall not discuss this further here.

\section{Extra experimental results}
\label{app:extra_results}

\subsection{Toy example}

\begin{figure}
	\centering
	\minput[pdf]{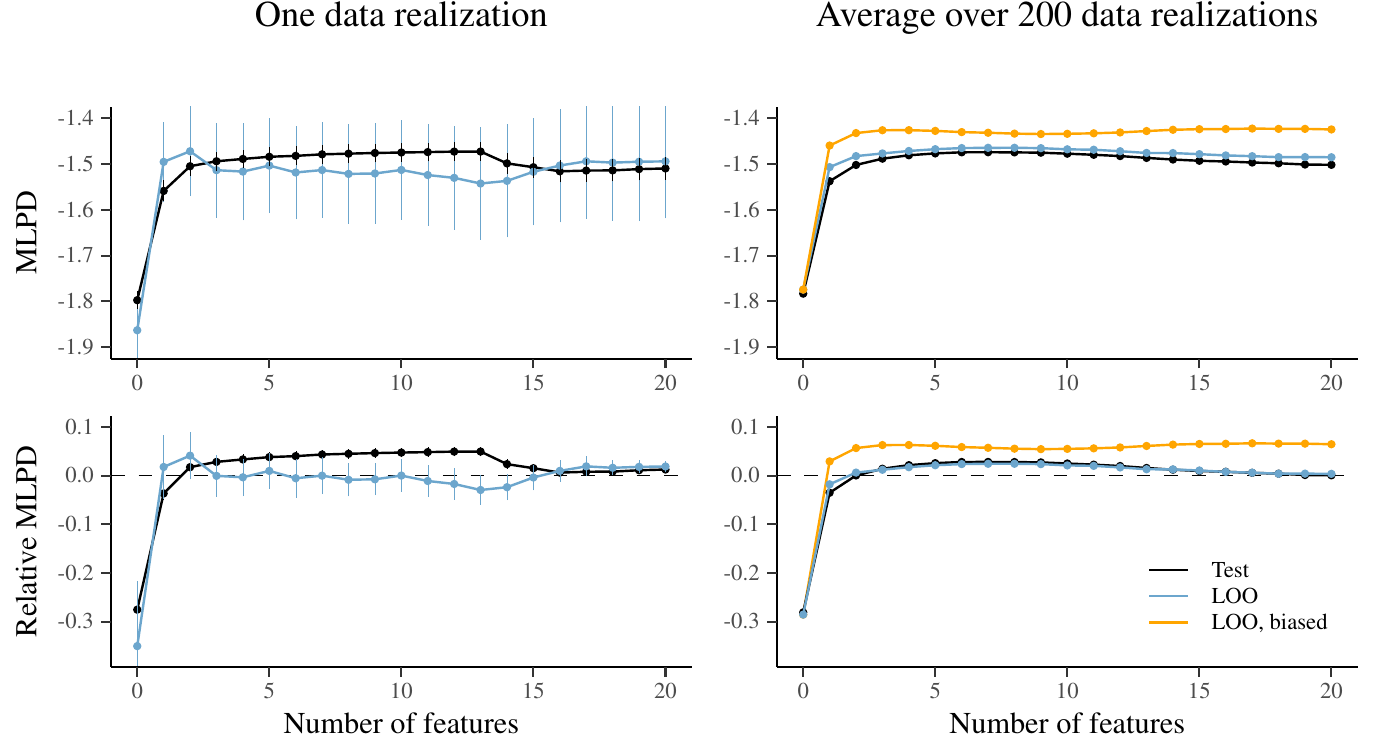}
	\caption{ {\it Toy example, projective selection:} The same as in Figure~\ref{fig:toy_assessment_rhs} but using the full model with Gaussian prior as the reference model. } 
	\label{fig:toy_assessment_g}
\end{figure}

Figure~\ref{fig:toy_assessment_g} shows the analogous results to Figure~\ref{fig:toy_assessment_rhs} but using the full model with Gaussian prior as the reference model.
The results are otherwise similar to those in Figure~\ref{fig:toy_assessment_rhs}, but here the submodels with 3 to 14 features achieve a slightly better generalization performance than the reference model.
This is simply due to the fact that the Gaussian prior is not the optimal choice in this particular case since it does not help to shrink the coefficients of the truly irrelevant features, and hence we can gain by removing those features.
This does not mean that the Gaussian prior would always be inappropriate even for very high-dimensional problems, see for instance the results for the microarray data sets in Section~\ref{sec:microarray_data} where the ridge regression performs very well (corresponds to maximum a posteriori solution with the Gaussian prior).

\subsection{Real world benchmarks}

Table~\ref{tab:nfeat} shows the average number of features selected by the different methods in the microarray examples (Sec.~\ref{sec:microarray_data}).
The projection methods clearly select the most parsimonious models. 

\begin{table}%
\centering
\abovetopsep=2pt
\begin{tabular}{ lrrrrrr }
\toprule
Method & Ovarian & Colon & Prostate & Leukemia & Glioma & Average \\
\midrule
Bayes SPC                                    & 633.0 & 881.0 &     7.7 &  1030.4 &   736.2&  657.7\\
Proj ref-1se                                 &   3.3 &   2.2 &     2.9 &     8.6 &    1.6 &   3.7\\
Proj ref-1se-reg                             &   7.2 &   3.9 &     3.8 &    11.0 &    7.2 &   6.6\\
Proj best-1se                                &   3.1 &   2.1 &     2.9 &    16.2 &    1.6 &   5.2\\
Proj best-1se-reg                            &   7.2 &   4.3 &     4.6 &    18.8 &    6.2 &   8.2\\
Lasso, $\lambda_\text{1se}$                &  11.7 &   5.1 &    12.7 &    15.3 &    9.2 &  10.8\\
Lasso, $\lambda_\text{best}$               &  18.6 &  11.1 &    23.6 &    23.2 &   25.0 &  20.3\\
Enet $\alpha=0.7$, $\lambda_\text{1se}$   &  22.5 &   8.1 &    23.2 &    25.9 &   23.7 &  20.7\\
Enet $\alpha=0.7$, $\lambda_\text{best}$  &  42.6 &  21.3 &    49.3 &    50.6 &   56.1 &  44.0\\
Enet $\alpha=0.3$, $\lambda_\text{1se}$   &  57.7 &  24.8 &    53.6 &    74.5 &   74.1 &  56.9\\
Enet $\alpha=0.3$, $\lambda_\text{best}$  & 114.3 &  64.2 &   124.8 &   188.0 &  187.7 & 135.8\\
Ridge, $\lambda_\text{1se}$                &1536 &2000 &  5966 &  7129 &22283 &7782.8\\
Ridge, $\lambda_\text{best}$               &1536 &2000 &  5966 &  7129 &22283 &7782.8\\
\bottomrule
\end{tabular}
\caption{Average number of features selected for the different methods in the microarray examples over the ten outer cross-validation folds. The last column denotes average over all data sets. The projection methods select by the most parsimonious models. The sparsest other method with similar predictive accuracy (Lasso with $\lambda_\tx{best}$, see Fig.~\ref{fig:realdata}) selects on average over twice as many features as the most dense projection (Proj best-1se-reg).}
\label{tab:nfeat}
\end{table}

\section{Proofs of the theoretical results} 
\label{app:proofs}

\subsection{Proof of Lemma~\ref{lemma:gain}}

First plug in the definitions~\eqref{eq:coeff_ls_and_ref} into the formula of the expected gain~\eqref{eq:gain} and expand
\begin{align*}
	nG &= n\left( \Delta(\vs{\hat \beta}) - \Delta(\vs \beta_*) \right) \\
	&= \mean[\vc{\ti y}]{ || \vc X \vs{\hat \beta}  - \vc{\ti y} ||^2 }
	- \mean[\vc{\ti y}]{ || \vc X \vs \beta_*  - \vc{\ti y} ||^2 } \\
	&= \vs{\hat \beta}^\tp \vc X^\tp \vc X \vs{\hat \beta} 
	- 2 \mean{\vc{\ti y}}^\tp \vc X \vs{\hat \beta} 
	+ \mean{\vc{\ti y}^\tp \vc{\ti y}} -
	\vs \beta_*^\tp \vc X^\tp \vc X \vs \beta_* 
	+ 2 \mean{\vc{\ti y}}^\tp \vc X \vs \beta_* 
	- \mean{\vc{\ti y}^\tp \vc{\ti y}} \\
	&= \vc y^\tp \vc X (\vc X^\tp \vc X)^{-1} \vc X^\tp \vc y 
	-2 \mean{\vc{\ti y}}^\tp \vc X (\vc X^\tp \vc X)^{-1} \vc X^\tp \vc y \\
	&\phantom{=} - \vs \mu_*^\tp \vc X (\vc X^\tp \vc X)^{-1} \vc X^\tp \vs \mu_* 
	+ 2 \mean{\vc{\ti y}}^\tp \vc X (\vc X^\tp \vc X)^{-1} \vc X^\tp \vs \mu_*.
\end{align*}
By plugging in $\mean{\vc{\ti y}} = \vs \mu$ and the definition $\vc P = \vc X (\vc X^\tp \vc X)^{-1} \vc X^\tp$, we get
\begin{align*}
n G	&= \vc y^\tp \vc P \vc y 
	-2 \vs \mu^\tp \vc P \vc y 
	- \vs \mu_*^\tp \vc P \vs \mu_* 
	+ 2 \vs \mu^\tp \vc P \vs \mu_*  \\
	&= \vc y^\tp \vc P \vc y 
	-2 \vs \mu^\tp \vc P \vc y 
	- \vs \mu_*^\tp \vc P \vs \mu_* 
	+ 2 \vs \mu^\tp \vc P \vs \mu_* 
	- \vs \mu^\tp \vc P \vs \mu
	+ \vs \mu^\tp \vc P \vs \mu  \\ 
	&= (\vc y - \vs \mu)^\tp \vc P (\vc y - \vs \mu)
	- (\vs \mu_* - \vs \mu)^\tp \vc P (\vs \mu_* - \vs \mu)	\\ 
	&= ||\vc y - \vs \mu||_{\vc P}^2
	- ||\vs \mu_* - \vs \mu||_{\vc P}^2.
\end{align*}
Hence  $G = \frac{1}{n} \left( ||\vc y - \vs \mu||_{\vc P}^2 - ||\vs \mu_* - \vs \mu||_{\vc P}^2 \right)$ .

\subsection{Proof of Theorem~\ref{thm:gain}}

By Lemma~\ref{lemma:gain}, $G = \frac{1}{n} \left( ||\vs \varepsilon||_{\vc P}^2 - ||\vs \mu_* - \vs \mu||_{\vc P}^2 \right)$, where $\vs \varepsilon = \vc y - \vs \mu$. 
Taking the expectation of $G$ with respect to the $\vs \varepsilon$ as well as the randomness in the reference fit $\vs \mu_*$ yields
\begin{align*}
	\mean{G} 
	&= \frac{1}{n}\left( \mean{\vs \varepsilon^\tp \vc P \vs \varepsilon} 
	-  \mean{(\vs \mu_* - \vs \mu)^\tp \vc P (\vs \mu_* - \vs \mu)} \right)\\
	&= \frac{1}{n} \left( \tr{\vc P \cov{\vs \varepsilon}} - \tr{\vc P \cov{\vs \mu_* - \vs \mu}} 
	  - \mean{\vs \mu_*-\vs \mu}^\tp \vc P \mean{\vs \mu_* - \vs \mu} \right) \\
	&= \frac{1}{n} \left( \sigma^2 \tr{\vc P} - \tr{\vc P \vc K} - \vc b^\tp \vc P \vc b \right).
\end{align*}
Now we have
\begin{align*}
	\tr{\vc P} 
	= \tr{\vc X (\vc X^\tp \vc X)^{-1} \vc X^\tp}
	= \tr{\vc X^\tp \vc X (\vc X^\tp \vc X)^{-1}}
	= \tr{\vc I_p}
	= p,
\end{align*}
so the expected gain simplifies to
\begin{align}
	\mean{G} = \frac{1}{n} \left( \sigma^2 p - \tr{\vc P \vc K} - ||\vc b||_{\vc P}^2 \right).
\label{eq:gain_thm}
\end{align}

\subsection{Proof of Corollary~\ref{clry:uncor_err}}

If the errors in the reference model are uncorrelated, that is, $\vc K = \sigma_{\mu_*}^2 \vc I$, the expected gain~\eqref{eq:gain_thm} reduces to
\begin{align*}
	\mean{G} &= \frac{1}{n} \left( \sigma^2 p - \sigma_{\mu_*}^2 p  -  ||\vc b||_{\vc P}^2 \right) \\
			&= \frac{p}{n} \left( \sigma^2  - \sigma_{\mu_*}^2   - \frac{1}{p} ||\vc b||_{\vc P}^2 \right).
\end{align*}

\end{appendices}

\bibliography{references}

\begin{thebibliography}{70}
\providecommand{\natexlab}[1]{#1}
\providecommand{\url}[1]{\texttt{#1}}
\expandafter\ifx\csname urlstyle\endcsname\relax
  \providecommand{\doi}[1]{doi: #1}\else
  \providecommand{\doi}{doi: \begingroup \urlstyle{rm}\Url}\fi

\bibitem[Ambroise and McLachlan(2002)]{ambroise2002}
Christophe Ambroise and Geoffrey~J. McLachlan.
\newblock Selection bias in gene extraction on the basis of microarray
  gene-expression data.
\newblock \emph{Proceedings of the National Academy of Sciences}, 99\penalty0
  (10):\penalty0 6562--6566, 2002.
\newblock \doi{10.1073/pnas.102102699}.

\bibitem[Armagan et~al.(2011)Armagan, Clyde, and Dunson]{armagan2011}
Artin Armagan, Merlise Clyde, and David~B Dunson.
\newblock Generalized beta mixtures of {G}aussians.
\newblock In J.~Shawe-Taylor, R.~S. Zemel, P.~L. Bartlett, F.~Pereira, and
  K.~Q. Weinberger, editors, \emph{Advances in Neural Information Processing
  Systems 24}, pages 523--531. Curran Associates, Inc., 2011.

\bibitem[Bair et~al.(2006)Bair, Hastie, Paul, and Tibshirani]{bair2006}
Eric Bair, Trevor Hastie, Debashis Paul, and Robert Tibshirani.
\newblock Prediction by supervised principal components.
\newblock \emph{{Journal of the American Statistical Association}},
  101\penalty0 (473):\penalty0 119--137, 2006.

\bibitem[Barbieri and Berger(2004)]{barbieri2004}
Maria~Maddalena Barbieri and James~O. Berger.
\newblock Optimal predictive model selection.
\newblock \emph{{The Annals of Statistics}}, 32\penalty0 (3):\penalty0
  870--897, 2004.
\newblock ISSN 0090-5364.
\newblock \doi{10.1214/009053604000000238}.

\bibitem[Bernardo and Ju{\'a}rez(2003)]{bernardo2003}
Jos\'{e}~M. Bernardo and Miguel~A. Ju{\'a}rez.
\newblock Intrinsic estimation.
\newblock In J.~M. Bernardo, M.~J. Bayarri, J.~O. Berger, A.~P. Dawid,
  D.~Heckerman, A.~F.~M. Smith, and M.~West, editors, \emph{Bayesian Statistics
  7}, pages 465--476. Oxford University Press, 2003.

\bibitem[Bernardo and Smith(1994)]{bernardo1994book}
Jos\'{e}~M. Bernardo and Adrian F.~M. Smith.
\newblock \emph{Bayesian Theory}.
\newblock John Wiley \& Sons, 1994.
\newblock \doi{10.1002/9780470316870}.

\bibitem[Bhadra et~al.(2017)Bhadra, Datta, Polson, and Willard]{bhadra2017}
Anindya Bhadra, Jyotishka Datta, Nicholas~G. Polson, and Brandon Willard.
\newblock The horseshoe$+$ estimator of ultra-sparse signals.
\newblock \emph{{Bayesian Analysis}}, 12\penalty0 (4):\penalty0 1105--1131,
  2017.
\newblock \doi{10.1214/16-BA1028}.

\bibitem[Bhattacharya et~al.(2015)Bhattacharya, Pati, Pillai, and
  Dunson]{bhattacharya2015}
Anirban Bhattacharya, Debdeep Pati, Natesh~S. Pillai, and David~B. Dunson.
\newblock Dirichlet-{L}aplace priors for optimal shrinkage.
\newblock \emph{{Journal of the American Statistical Association}},
  110\penalty0 (512):\penalty0 1479--1490, 2015.
\newblock ISSN 0162-1459.
\newblock \doi{10.1080/01621459.2014.960967}.

\bibitem[Breiman(1995)]{breiman1995}
Leo Breiman.
\newblock Better subset regression using the nonnegative garrote.
\newblock \emph{Technometrics}, 37\penalty0 (4):\penalty0 373--384, 1995.

\bibitem[Bucil\v{a} et~al.(2006)Bucil\v{a}, Caruana, and
  Niculescu-Mizil]{bucila2006}
Cristian Bucil\v{a}, Rich Caruana, and Alexandru Niculescu-Mizil.
\newblock Model compression.
\newblock In \emph{Proceedings of the 12th ACM SIGKDD International Conference
  on Knowledge Discovery and Data Mining}, KDD '06, pages 535--541. ACM, 2006.

\bibitem[B\"{u}rkner(2017)]{burkner2017}
Paul-Christian B\"{u}rkner.
\newblock brms: An {R} package for {B}ayesian multilevel models using {S}tan.
\newblock \emph{Journal of Statistical Software, Articles}, 80\penalty0
  (1):\penalty0 1--28, 2017.
\newblock \doi{10.18637/jss.v080.i01}.

\bibitem[Candes and Tao(2007)]{candes2007}
Emmanuel Candes and Terence Tao.
\newblock The {D}antzig selector: statistical estimation when $p$ is much
  larger than $n$.
\newblock \emph{{The Annals of Statistics}}, 35\penalty0 (6):\penalty0
  2313--2351, 2007.

\bibitem[Carvalho et~al.(2009)Carvalho, Polson, and Scott]{carvalho2009}
Carlos~M. Carvalho, Nicholas~G. Polson, and James~G. Scott.
\newblock Handling sparsity via the horseshoe.
\newblock In David van Dyk and Max Welling, editors, \emph{Proceedings of the
  12th International Conference on Artificial Intelligence and Statistics},
  volume~5 of \emph{Proceedings of Machine Learning Research}, pages 73--80.
  PMLR, 2009.

\bibitem[Carvalho et~al.(2010)Carvalho, Polson, and Scott]{carvalho2010}
Carlos~M. Carvalho, Nicholas~G. Polson, and James~G. Scott.
\newblock The horseshoe estimator for sparse signals.
\newblock \emph{Biometrika}, 97\penalty0 (2):\penalty0 465--480, 2010.
\newblock \doi{10.1093/biomet/asq017}.

\bibitem[Cawley and Talbot(2010)]{cawley2010}
Gavin~C. Cawley and Nicola L.~C. Talbot.
\newblock On over-fitting in model selection and subsequent selection bias in
  performance evaluation.
\newblock \emph{Journal of Machine Learning Research}, 11:\penalty0 2079--2107,
  2010.

\bibitem[Dupuis and Robert(2003)]{dupuis2003}
J{\'e}rome~A. Dupuis and Christian~P. Robert.
\newblock Variable selection in qualitative models via an entropic explanatory
  power.
\newblock \emph{Journal of Statistical Planning and Inference}, 111\penalty0
  (1-2):\penalty0 77--94, 2003.
\newblock ISSN 0378-3758.
\newblock \doi{10.1016/S0378-3758(02)00286-0}.

\bibitem[Efron(2010)]{efron2010book}
Bradley Efron.
\newblock \emph{Large-scale inference}, volume~1 of \emph{Institute of
  Mathematical Statistics (IMS) Monographs}.
\newblock Cambridge University Press, Cambridge, 2010.
\newblock \doi{10.1017/CBO9780511761362}.
\newblock Empirical {B}ayes methods for estimation, testing, and prediction.

\bibitem[Efron et~al.(2004)Efron, Hastie, Johnstone, and Tibshirani]{efron2004}
Bradley Efron, Trevor Hastie, Iain Johnstone, and Robert Tibshirani.
\newblock Least angle regression.
\newblock \emph{{The Annals of Statistics}}, 32\penalty0 (2):\penalty0
  407--499, 2004.
\newblock ISSN 0090-5364.
\newblock \doi{10.1214/009053604000000067}.

\bibitem[Fan and Li(2001)]{fan2001}
J.~Fan and R.~Li.
\newblock Variable selection via nonconcave penalized likelihood and its oracle
  properties.
\newblock \emph{{Journal of the American Statistical Association}}, 96\penalty0
  (456):\penalty0 1348--1360, 2001.

\bibitem[Fan and Lv(2008)]{fan2008}
J.~Fan and J.~Lv.
\newblock Sure independence screening for ultrahigh dimensional feature space.
\newblock \emph{{Journal of the Royal Statistical Society. Series B
  (Methodological)}}, 70\penalty0 (5):\penalty0 849--911, 2008.

\bibitem[Friedman et~al.(2010)Friedman, Hastie, and Tibshirani]{friedman2010}
Jerome Friedman, Trevor Hastie, and Robert Tibshirani.
\newblock Regularization paths for generalized linear models via coordinate
  descent.
\newblock \emph{Journal of Statistical Software}, 33\penalty0 (1), 2010.

\bibitem[Gabry et~al.(2018)Gabry, Simpson, Vehtari, Betancourt, and
  Gelman]{gabry2018}
Jonah Gabry, Daniel Simpson, Aki Vehtari, Michael Betancourt, and Andrew
  Gelman.
\newblock Visualization in bayesian workflow.
\newblock \emph{{Journal of the Royal Statistical Society. Series A}}, Accepted
  for publication, 2018.

\bibitem[Gelman et~al.(2013)Gelman, Carlin, Stern, Dunson, Vehtari, and
  Rubin]{gelman2013book}
Andrew Gelman, John~B. Carlin, Hal~S. Stern, David~B. Dunson, Aki Vehtari, and
  Donald~B. Rubin.
\newblock \emph{Bayesian Data Analysis}.
\newblock Chapman \& Hall, {Third} edition, 2013.

\bibitem[George and McCulloch(1993)]{george1993}
Edward~I. George and Robert~E. McCulloch.
\newblock Variable selection via {G}ibbs sampling.
\newblock \emph{{Journal of the American Statistical Association}}, 88\penalty0
  (423):\penalty0 881--889, 1993.

\bibitem[Goodrich et~al.(2018)Goodrich, Gabry, Ali, and Brilleman]{rstanarm}
Ben Goodrich, Jonah Gabry, Imad Ali, and Sam Brilleman.
\newblock rstanarm: {B}ayesian applied regression modeling via {S}tan, 2018.
\newblock URL \url{http://mc-stan.org/}.
\newblock R package version 2.17.4.

\bibitem[Goutis and Robert(1998)]{goutis1998}
Constantinos Goutis and Christian~P. Robert.
\newblock Model choice in generalised linear models: {A} {B}ayesian approach
  via {K}ullback--{L}eibler projections.
\newblock \emph{Biometrika}, 85\penalty0 (1):\penalty0 29--37, 1998.
\newblock ISSN 0006-3444.
\newblock \doi{10.1093/biomet/85.1.29}.

\bibitem[Hahn and Carvalho(2015)]{hahn2015}
P.~Richard Hahn and Carlos~M. Carvalho.
\newblock Decoupling shrinkage and selection in {B}ayesian linear models: a
  posterior summary perspective.
\newblock \emph{Journal of the American Statistical Association}, 110\penalty0
  (509):\penalty0 435--448, 2015.
\newblock ISSN 0162-1459.
\newblock \doi{10.1080/01621459.2014.993077}.

\bibitem[Hastie et~al.(2009)Hastie, Tibshirani, and Friedman]{hastie2009book}
Trevor Hastie, Robert Tibshirani, and Jerome Friedman.
\newblock \emph{The Elements of Statistical Learning}.
\newblock Springer-Verlag, {Second} edition, 2009.

\bibitem[Hastie et~al.(2015)Hastie, Tibshirani, and Wainwright]{hastie2015book}
Trevor Hastie, Robert Tibshirani, and Martin Wainwright.
\newblock \emph{Statistical learning with sparsity: the {L}asso and
  generalizations}.
\newblock Chapman \& Hall, 2015.

\bibitem[Hern{\'a}ndez-Lobato et~al.(2010)Hern{\'a}ndez-Lobato,
  Hern{\'a}ndez-Lobato, and Su{\'a}rez]{hernandezlobato2010}
Daniel Hern{\'a}ndez-Lobato, Jos{\'e}~Miguel Hern{\'a}ndez-Lobato, and Alberto
  Su{\'a}rez.
\newblock Expectation propagation for microarray data classification.
\newblock \emph{Pattern Recognition Letters}, 31\penalty0 (12):\penalty0
  1618--1626, 2010.

\bibitem[Hinton et~al.(2015)Hinton, Vinyals, and Dean]{hinton2015}
Geoffrey Hinton, Oriol Vinyals, and Jeff Dean.
\newblock Distilling the knowledge in a neural network.
\newblock \emph{arXiv:1503.02531}, 2015.

\bibitem[Ishwaran and Rao(2005)]{ishwaran2005}
Hemant Ishwaran and J.~Sunil Rao.
\newblock Spike and slab variable selection: frequentist and {B}ayesian
  strategies.
\newblock \emph{{The Annals of Statistics}}, 33\penalty0 (2):\penalty0
  730--773, 2005.
\newblock \doi{10.1214/009053604000001147}.

\bibitem[Ishwaran et~al.(2010)Ishwaran, Kogalur, and Rao]{ishwaran2010}
Hemant Ishwaran, Udaya~B. Kogalur, and J.~Sunil Rao.
\newblock {spikeslab}: {P}rediction and variable selection using spike and slab
  regression.
\newblock \emph{The {R} Journal}, 2\penalty0 (2):\penalty0 68--73, 2010.

\bibitem[Jeffreys(1961)]{jeffreys1961book}
Harold Jeffreys.
\newblock \emph{Theory of Probability}.
\newblock Oxford University Press, 3rd edition, 1961.
\newblock (1st edition 1939).

\bibitem[Johnson and Rossell(2012)]{johnson2012}
Valen~E. Johnson and David Rossell.
\newblock Bayesian model selection in high-dimensional settings.
\newblock \emph{{Journal of the American Statistical Association}},
  107\penalty0 (498):\penalty0 649--660, 2012.
\newblock ISSN 0162-1459.
\newblock \doi{10.1080/01621459.2012.682536}.

\bibitem[Johnstone and Silverman(2004)]{johnstone2004}
Iain~M. Johnstone and Bernard~W. Silverman.
\newblock Needles and straw in haystacks: empirical {B}ayes estimates of
  possibly sparse sequences.
\newblock \emph{{The Annals of Statistics}}, 32\penalty0 (4):\penalty0
  1594--1649, 2004.
\newblock ISSN 0090-5364.
\newblock \doi{10.1214/009053604000000030}.
\newblock URL \url{http://dx.doi.org/10.1214/009053604000000030}.

\bibitem[Kass and Raftery(1995)]{kass1995}
Robert~E. Kass and Adrian~E. Raftery.
\newblock {B}ayes factors.
\newblock \emph{{Journal of the American Statistical Association}}, 90\penalty0
  (430):\penalty0 773--795, 1995.

\bibitem[Lee et~al.(2003)Lee, Sha, Dougherty, Vannucci, and Mallick]{lee2003}
Kyeong~Eun Lee, Naijun Sha, Edward~R Dougherty, Marina Vannucci, and Bani~K
  Mallick.
\newblock Gene selection: a {B}ayesian variable selection approach.
\newblock \emph{Bioinformatics}, 19\penalty0 (1):\penalty0 90--97, 2003.

\bibitem[Li et~al.(2002)Li, Campbell, and Tipping]{li2002}
Yi~Li, Colin Campbell, and Michael Tipping.
\newblock {B}ayesian automatic relevance determination algorithms for
  classifying gene expression data.
\newblock \emph{Bioinformatics}, 18\penalty0 (10):\penalty0 1332--1339, 2002.

\bibitem[Lindley(1968)]{lindley1968}
D.~V. Lindley.
\newblock The choice of variables in multiple regression.
\newblock \emph{{Journal of the Royal Statistical Society. Series B
  (Methodological)}}, 30:\penalty0 31--66, 1968.
\newblock ISSN 0035-9246.

\bibitem[McCullagh and Nelder(1989)]{mccullagh1989book}
P.~McCullagh and J.~A. Nelder.
\newblock \emph{Generalized linear models}.
\newblock Monographs on Statistics and Applied Probability. Chapman \& Hall,
  second edition, 1989.
\newblock ISBN 0-412-31760-5.
\newblock \doi{10.1007/978-1-4899-3242-6}.
\newblock URL \url{http://dx.doi.org/10.1007/978-1-4899-3242-6}.

\bibitem[Meinshausen(2007)]{meinshausen2007}
Nicolai Meinshausen.
\newblock Relaxed {L}asso.
\newblock \emph{{Computational Statistics \& Data Analysis}}, 52\penalty0
  (1):\penalty0 374--393, 2007.
\newblock ISSN 0167-9473.
\newblock \doi{10.1016/j.csda.2006.12.019}.

\bibitem[Narisetty and He(2014)]{narisetty2014}
Naveen~Naidu Narisetty and Xuming He.
\newblock {B}ayesian variable selection with shrinking and diffusing priors.
\newblock \emph{{The Annals of Statistics}}, 42\penalty0 (2):\penalty0
  789--817, 2014.
\newblock ISSN 0090-5364.
\newblock \doi{10.1214/14-AOS1207}.

\bibitem[Neal and Zhang(2006)]{neal2006}
Radford Neal and Jianguo Zhang.
\newblock High dimensional classification with {B}ayesian neural networks and
  {D}irichlet diffusion trees.
\newblock In Isabelle Guyon, Steve Gunn, Masoud Nikravesh, and Lotfi~A. Zadeh,
  editors, \emph{Feature Extraction, Foundations and Applications}, pages
  265--296. Springer, 2006.

\bibitem[Nott and Leng(2010)]{nott2010}
David~J. Nott and Chenlei Leng.
\newblock {B}ayesian projection approaches to variable selection in generalized
  linear models.
\newblock \emph{Computational Statistics and Data Analysis}, 54:\penalty0
  3227--3241, 2010.

\bibitem[Paul et~al.(2008)Paul, Bair, Hastie, and Tibshirani]{paul2008}
Debashis Paul, Eric Bair, Trevor Hastie, and Robert Tibshirani.
\newblock ``{P}reconditioning'' for feature selection and regression in
  high-dimensional problems.
\newblock \emph{{The Annals of Statistics}}, 36\penalty0 (4):\penalty0
  1595--1618, 2008.

\bibitem[Peltola(2018)]{peltola2018}
Tomi Peltola.
\newblock Local interpretable model-agnostic explanations of {B}ayesian
  predictive models via {K}ullback-{L}eibler projections.
\newblock In David~W. Aha, Trevor Darrell, Patrick Doherty, and Daniele
  Magazzeni, editors, \emph{Proceedings of the 2nd Workshop on Explainable
  Artificial Intelligence}, pages 114--118, 2018.

\bibitem[Peltola et~al.(2014)Peltola, Havulinna, Salomaa, and
  Vehtari]{peltola2014}
Tomi Peltola, Aki~S Havulinna, Veikko Salomaa, and Aki Vehtari.
\newblock Hierarchical {B}ayesian survival analysis and projective covariate
  selection in cardiovascular event risk prediction.
\newblock In \emph{Proceedings of the 11th {UAI} {B}ayesian Modeling
  Applications Workshop}, volume 1218 of \emph{CEUR Workshop Proceedings},
  pages 79--88, 2014.

\bibitem[Piironen and Vehtari(2016)]{piironen2016}
Juho Piironen and Aki Vehtari.
\newblock Projection predictive model selection for {G}aussian processes.
\newblock In \emph{2016 IEEE 26th International Workshop on Machine Learning
  for Signal Processing (MLSP)}, pages 1--6. IEEE, 2016.

\bibitem[Piironen and Vehtari(2017{\natexlab{a}})]{piironen2017a}
Juho Piironen and Aki Vehtari.
\newblock Comparison of {B}ayesian predictive methods for model selection.
\newblock \emph{Statistics and Computing}, 27\penalty0 (3):\penalty0 711--735,
  2017{\natexlab{a}}.
\newblock \doi{10.1007/s11222-016-9649-y}.

\bibitem[Piironen and Vehtari(2017{\natexlab{b}})]{piironen2017b}
Juho Piironen and Aki Vehtari.
\newblock On the hyperprior choice for the global shrinkage parameter in the
  horseshoe prior.
\newblock In Aarti Singh and Jerry Zhu, editors, \emph{Proceedings of the 20th
  International Conference on Artificial Intelligence and Statistics},
  volume~54 of \emph{Proceedings of Machine Learning Research}, pages 905--913.
  PMLR, 2017{\natexlab{b}}.
\newblock URL \url{http://proceedings.mlr.press/v54/piironen17a.html}.

\bibitem[Piironen and Vehtari(2017{\natexlab{c}})]{piironen2017c}
Juho Piironen and Aki Vehtari.
\newblock Sparsity information and regularization in the horseshoe and other
  shrinkage priors.
\newblock \emph{{Electronic Journal of Statistics}}, 11\penalty0 (2):\penalty0
  5018--5051, 2017{\natexlab{c}}.
\newblock \doi{10.1214/17-EJS1337SI}.

\bibitem[Piironen and Vehtari(2018)]{piironen2018}
Juho Piironen and Aki Vehtari.
\newblock Iterative supervised principal components.
\newblock In Amos Storkey and Fernando Perez-Cruz, editors, \emph{Proceedings
  of the 21st International Conference on Artificial Intelligence and
  Statistics}, volume~84 of \emph{Proceedings of Machine Learning Research},
  pages 106--114. PMLR, 2018.

\bibitem[Polson and Scott(2011)]{polson2011}
Nicholas~G. Polson and James~G. Scott.
\newblock Shrink globally, act locally: sparse {B}ayesian regularization and
  prediction.
\newblock In J.~M. Bernardo, M.~J. Bayarri, J.~O. Berger, A.~P. Dawid,
  D.~Heckerman, A.~F.~M. Smith, and M.~West, editors, \emph{Bayesian statistics
  9}, pages 501--538. Oxford University Press, Oxford, 2011.
\newblock \doi{10.1093/acprof:oso/9780199694587.003.0017}.

\bibitem[Raftery et~al.(1997)Raftery, Madigan, and Hoeting]{raftery1997}
Adrian~E. Raftery, David Madigan, and Jennifer~A. Hoeting.
\newblock {B}ayesian model averaging for linear regression models.
\newblock \emph{{Journal of the American Statistical Association}}, 92\penalty0
  (437):\penalty0 179--191, 1997.
\newblock ISSN 0162-1459.
\newblock \doi{10.2307/2291462}.

\bibitem[Reid et~al.(2016)Reid, Tibshirani, and Friedman]{reid2016}
Stephen Reid, Robert Tibshirani, and Jerome Friedman.
\newblock A study of error variance estimation in {L}asso regression.
\newblock \emph{Statistica Sinica}, 26\penalty0 (1):\penalty0 35--67, 2016.
\newblock ISSN 1017-0405.

\bibitem[Reunanen(2003)]{reunanen2003}
Juha Reunanen.
\newblock Overfitting in making comparisons between variable selection methods.
\newblock \emph{Journal of Machine Learning Research}, 3:\penalty0 1371--1382,
  2003.

\bibitem[Ribeiro et~al.(2016)Ribeiro, Singh, and Guestrin]{ribeiro2016}
Marco~Tulio Ribeiro, Sameer Singh, and Carlos Guestrin.
\newblock ``{W}hy should {I} trust you?'' {E}xplaining the predictions of any
  classifier.
\newblock In \emph{Proceedings of the 22nd ACM SIGKDD International Conference
  on Knowledge Discovery and Data Mining}, KDD '16, pages 1135--1144. ACM,
  2016.

\bibitem[Snelson and Ghahramani(2005)]{snelson2005}
Edward Snelson and Zoubin Ghahramani.
\newblock Compact approximations to {B}ayesian predictive distributions.
\newblock In \emph{Proceedings of the 22nd International Conference on Machine
  Learning}, ICML '05, pages 840--847. ACM, 2005.
\newblock \doi{10.1145/1102351.1102457}.

\bibitem[{S}tan~{D}evelopment {T}eam(2018)]{stan_manual}
{S}tan~{D}evelopment {T}eam.
\newblock {S}tan modeling language users guide and reference manual, version
  2.18.0, 2018.
\newblock URL \url{http://mc-stan.org}.

\bibitem[Tibshirani(1996)]{tibshirani1996}
Robert Tibshirani.
\newblock Regression shrinkage and selection via the {L}asso.
\newblock \emph{{Journal of the Royal Statistical Society. Series B
  (Methodological)}}, 58\penalty0 (1):\penalty0 267--288, 1996.

\bibitem[Tran et~al.(2012)Tran, Nott, and Leng]{tran2012}
Minh-Ngoc Tran, David~J. Nott, and Chenlei Leng.
\newblock The predictive {L}asso.
\newblock \emph{Statistics and Computing}, 22\penalty0 (5):\penalty0
  1069--1084, 2012.
\newblock ISSN 0960-3174.
\newblock \doi{10.1007/s11222-011-9279-3}.

\bibitem[van~der Pas et~al.(2014)van~der Pas, Kleijn, and van~der
  Vaart]{vanderpas2014}
S.~L. van~der Pas, B.~J.~K. Kleijn, and A.~W. van~der Vaart.
\newblock The horseshoe estimator: posterior concentration around nearly black
  vectors.
\newblock \emph{{Electronic Journal of Statistics}}, 8\penalty0 (2):\penalty0
  2585--2618, 2014.
\newblock ISSN 1935-7524.
\newblock \doi{10.1214/14-EJS962}.

\bibitem[van~der Pas et~al.(2017)van~der Pas, Szabó, and van~der
  Vaart]{vanderpas2017}
Stéphanie van~der Pas, Botond Szabó, and Aad van~der Vaart.
\newblock Uncertainty quantification for the horseshoe.
\newblock \emph{{Bayesian Analysis}}, 12\penalty0 (4):\penalty0 1221--1274,
  2017.
\newblock \doi{10.1214/17-BA1065}.

\bibitem[Vehtari and Ojanen(2012)]{vehtari2012}
Aki Vehtari and Janne Ojanen.
\newblock A survey of {B}ayesian predictive methods for model assessment,
  selection and comparison.
\newblock \emph{Statistics Surveys}, 6:\penalty0 142--228, 2012.
\newblock \doi{10.1214/12-SS102}.

\bibitem[Vehtari et~al.(2017{\natexlab{a}})Vehtari, Gelman, and
  Gabry]{vehtari2017a}
Aki Vehtari, Andrew Gelman, and Jonah Gabry.
\newblock Pareto smoothed importance sampling.
\newblock \emph{arXiv:1507.02646v5}, 2017{\natexlab{a}}.

\bibitem[Vehtari et~al.(2017{\natexlab{b}})Vehtari, Gelman, and
  Gabry]{vehtari2017b}
Aki Vehtari, Andrew Gelman, and Jonah Gabry.
\newblock Practical {B}ayesian model evaluation using leave-one-out
  cross-validation and {WAIC}.
\newblock \emph{Statistics and Computing}, 27\penalty0 (5):\penalty0
  1413--1432, 2017{\natexlab{b}}.
\newblock ISSN 0960-3174.
\newblock \doi{10.1007/s11222-016-9696-4}.

\bibitem[Yao et~al.(2018)Yao, Vehtari, Simpson, and Gelman]{yao2018}
Yuling Yao, Aki Vehtari, Daniel Simpson, and Andrew Gelman.
\newblock Using stacking to average {B}ayesian predictive distributions (with
  discussion).
\newblock \emph{{Bayesian Analysis}}, 13\penalty0 (3):\penalty0 917--1003, 09
  2018.
\newblock \doi{10.1214/17-BA1091}.

\bibitem[Zou(2006)]{zou2006}
Hui Zou.
\newblock The adaptive {L}asso and its oracle properties.
\newblock \emph{{Journal of the American Statistical Association}},
  101\penalty0 (476):\penalty0 1418--1429, 2006.
\newblock ISSN 0162-1459.
\newblock \doi{10.1198/016214506000000735}.

\bibitem[Zou and Hastie(2005)]{zou2005}
Hui Zou and Trevor Hastie.
\newblock Regularization and variable selection via the elastic net.
\newblock \emph{{Journal of the Royal Statistical Society. Series B
  (Methodological)}}, 67\penalty0 (2):\penalty0 301--320, 2005.
\newblock ISSN 1369-7412.
\newblock \doi{10.1111/j.1467-9868.2005.00503.x}.

\end{thebibliography}

\end{document}